\documentclass[twoside]{article}
\usepackage{ecj,palatino}
\usepackage{natbib}
\usepackage{booktabs} 
\usepackage{epsfig}
\usepackage{multirow}
\usepackage{subcaption}
\usepackage{booktabs}
\usepackage{ifthen}

\usepackage{wrapfig}
\graphicspath{{./fig/}{./fig/median/}{./fig/Ranking/}{./fig/unconstrained/}{./fig/Resampling/}{./fig/BCH/}}

\usepackage{color}
\usepackage[table,xcdraw]{xcolor}

\newcommand{\markupdraft}[2]{
  \ifthenelse{\equal{#1}{display}}{#2}{}
  \ifthenelse{\equal{#1}{color}}{\color{#2}}{}
}

\newcommand{\newcolored}[3][]{{\markupdraft{color}{#2}#3}
  \ifthenelse{\equal{#1}{}}{}{\markupdraft{display}{{\color{yellow!70!black}[#1]}}}}
\newcommand{\del}[2][]{{\markupdraft{display}{{\color{orange}[removed: ``#2''[#1]]}}}} 
\newcommand{\ddel}[2][]{{\markupdraft{display}{{\color{orange}[removed: ``#2''[#1]]}}}} 
\newcommand{\new}[2][]{\newcolored[#1]{blue}{#2}}
\newcommand{\nnew}[2][]{\newcolored[#1]{blue}{#2}}

\renewcommand{\del}[2]{}
\renewcommand{\new}[2][]{\newcolored[#1]{black}{#2}}
\renewcommand{\nnew}[2][]{\newcolored[#1]{black}{#2}}
\renewcommand{\ddel}[2]{}

\usepackage[T1]{fontenc}
\usepackage{latexsym}
\usepackage{amsmath,amssymb,amsthm} 
\usepackage{mathtools}       
\usepackage{stmaryrd}        
\newcommand{\bm}[1]{\boldsymbol{#1}}
\usepackage{braket}
\usepackage{dsfont}
\usepackage{empheq}

\providecommand{\argmin}{\operatornamewithlimits{argmin}} 
\DeclareMathOperator{\diag}{diag} 
\providecommand{\R}{\mathbb{R}} 
\providecommand{\Rp}{\R^+}
\providecommand{\E}{\mathbb{E}} 
\providecommand{\T}{\mathrm{T}} 
\renewcommand{\geq}{\geqslant} 
\renewcommand{\leq}{\leqslant} 
\DeclarePairedDelimiterX{\inner}[2]{\langle}{\rangle}{#1, #2}
\DeclarePairedDelimiter{\norm}{\lVert}{\rVert}
\DeclarePairedDelimiter{\abs}{\lvert}{\rvert}

\newcommand{\mueff}{\mu_\text{eff}}

\newcommand{\xmean}{{\bm{m}}}
\newcommand{\cov}{\bm{C}}
\newcommand{\dist}{\mathcal{N}(\xmean,\sigma^2 \cov)}
\renewcommand{\epsilon}{\varepsilon}
\newcommand{\eps}{\epsilon}

\providecommand{\NN}{\mathcal{N}}
\providecommand{\CC}{\cov}
\providecommand{\mm}{\xmean}
\providecommand{\xx}{\bm{x}}
\providecommand{\yy}{\bm{y}}
\providecommand{\zz}{\bm{z}}
\providecommand{\pc}{\bm{p}_c}
\providecommand{\ps}{\bm{p}_\sigma}
\providecommand{\cc}{c_\mathrm{c}}
\providecommand{\cs}{c_\mathrm{\sigma}}

\providecommand{\cone}{c_1}
\providecommand{\cmu}{c_\mu}
\providecommand{\ds}{d_\mathrm{\sigma}}
\providecommand{\ww}{\mathrm{w}}
\providecommand{\muw}{\mu_{\ww}}

\providecommand{\gs}{\gamma_\sigma}
\providecommand{\gc}{\gamma_\mathrm{c}}

\providecommand{\hsig}{h_\sigma}

\providecommand{\dy}{\langle \yy \rangle_\ww}

\newtheorem{theorem}{Theorem}
\def\eq#1{\eqref{#1}}

\def\ite#1{{(#1)}}

\newcommand{\HH}[0]{\bm{H}}

\newcommand{\g}{\bm{g}}

\newcommand{\step}[1]{{\textbf{\textsc{Step} #1}}}
\newcommand{\A}[0]{\bm{A}}
\newcommand{\bb}[0]{\bm{b}}
\newcommand{\lb}[0]{\texttt{LB}}
\newcommand{\ub}[0]{\texttt{UB}}
\newcommand{\xfeas}[0]{\xx^\mathrm{feas}}

\newcommand{\Rf}[0]{\mathrm{R}_f}
\newcommand{\RT}[0]{\mathrm{R}_T}
\newcommand{\Rg}[0]{\mathrm{R}_g}
\newcommand{\gsig}[0]{g_{\Sigma}}
\newcommand{\Cinv}[0]{{\bm{\Sigma}^{-1}}}
\newcommand{\fsph}[0]{f_\mathrm{sph}}
\newcommand{\fell}[0]{f_\mathrm{ell}}
\newcommand{\frotell}[0]{f_\mathrm{rotell}}

\newcommand{\rfeas}[0]{r_\mathrm{feas}}

\newcommand{\1}[1]{\mathds{1}_{\left\{ #1 \right\}}}
\newcommand{\vj}[1][j]{\bm{v}_{#1}}

\newcommand{\x}[0]{\bm{x}}

\newcommand{\snf}[0]{{\bar{\sigma}^*}}

\newcommand{\dm}[0]{d_{\mm}}

\newcommand{\sgn}[0]{\mathtt{sgn}}
\newcommand{\criterion}[0]{\| \mm - \xx^* \|^2_{\HH}}
\newcommand{\Prot}{\bm{P}_\mathrm{rot}}
\newcommand{\Pillrot}{\bm{P}_\mathrm{illrot}}
\newcommand{\BOX}{\texttt{Box}}
\newcommand{\rotBox}{\texttt{rotBox}}
\newcommand{\illrotBox}{\texttt{illrotBox}}
\newcommand{\J}[0]{\mathcal{J}}

\newcommand{\I}[0]{{\bm{I}_n}}

\newcommand{\cact}{{c_\mathrm{act}}}

\usepackage[colorlinks=false,pdfborder={0 0 0}]{hyperref}
\usepackage{cleveref}
\crefname{equation}{Eq.}{Eqs.}
\Crefname{equation}{Equation}{Equations}
\crefname{table}{table}{tables}
\Crefname{table}{Table}{Tables}

\crefname{figure}{Fig.}{Figs.}
\Crefname{figure}{Figure}{Figures}

\begin{document}
\ecjHeader{x}{x}{xxx-xxx}{201X}{Adaptive Ranking-based Constraint Handling}{N. Sakamoto and Y. Akimoto}
\title{\bf Adaptive Ranking-based Constraint Handling for Explicitly Constrained
  Black-Box Optimization}

\author{
  \name{\bf Naoki Sakamoto} \hfill \addr{naoki@bbo.cs.tsukuba.ac.jp}\\
\addr{Graduate School of Systems and Information Engineering, University of Tsukuba
  and RIKEN Center for Advanced Intelligence Project}
\AND \name{\bf Youhei Akimoto} \hfill \addr{akimoto@cs.tsukuba.ac.jp}\\
\addr{Faculty of Engineering, Information and Systems, University of Tsukuba and RIKEN
  Center for Advanced Intelligence Project}
}

\maketitle
\begin{abstract}
  We propose a novel constraint-handling technique for the covariance matrix
adaptation evolution strategy (CMA-ES). The proposed technique is aimed at solving
explicitly constrained black-box continuous optimization problems, in which the
explicit constraint is a constraint whereby the computational time for the
constraint violation and its (numerical) gradient are negligible compared to that
for the objective function. This method is designed to realize two invariance
properties: invariance to the affine transformation of the search space, and
invariance to the increasing transformation of the objective and constraint
functions.  The CMA-ES is designed to possess these properties for handling
difficulties that appear in\ddel{ constrained}{} black-box optimization problems,
such as non-separability, ill-conditioning, ruggedness, and the different orders
of magnitude in the objective\ddel{ and constraint functions}{}.  The proposed
constraint-handling technique (CHT), known as ARCH, modifies the underlying CMA-ES
only in terms of the ranking of the candidate solutions. It employs a repair
operator and an adaptive ranking aggregation strategy to compute the ranking.
\ddel{The effectiveness of the proposed technique lies in not only the
abovementioned invariance properties, but also its ability to handle nonlinear
constraints.}{}%
We developed test problems to evaluate the effects of the
invariance properties, and \ddel{empirically evaluated the performance of ARCH
from an invariance perspective}{}\nnew{performed experiments to empirically verify
the invariance of the algorithm}{}. We compared the proposed method with other
CHTs on the CEC 2006 constrained optimization benchmark suite to demonstrate its
efficacy. Empirical studies reveal that ARCH is able to exploit the explicitness
of the constraint functions effectively, sometimes even more efficiently than an
existing box-constraint handling technique on box-constrained problems, while
exhibiting the invariance properties. Moreover, ARCH overwhelmingly outperforms
CHTs by not exploiting the explicit constraints in terms of the number of
objective function calls.
\end{abstract}



\section{Introduction}
\label{sec:introduction}

We consider explicitly constrained black-box continuous minimization problems defined
as
\begin{align}
  \label{eq:orig-prob} \argmin_{\xx \in \R^n} \ f(\xx) \enspace \text{subject to}
\enspace g_j(\xx) \leq 0,\ \forall j = 1, \dots, m \enspace,
\end{align}%
where $f: \R^n \to \R$ is the objective function and $g_j: \R^n \to \R$
($j = 1, \dots, m$) are the constraint functions. The equality constraints $h_k(\xx) =
0, \ \forall k = 1, \dots, l$ are assumed to be transformed into inequality
constraints $g_k(\xx) = \abs{h_k(\xx)} - \epsilon_\mathrm{eq}$, with a numerical
tolerance $\epsilon_\mathrm{eq} > 0$. In the black-box optimization scenario, the
evaluation of the objective function $f$ often requires computationally expensive
simulation, while the \emph{explicit} constraints can be computed independently of
$f$-calls, and their computational cost is significantly lower than that of $f$-calls.
These can be presented in a relatively simple mathematical expression, and their
gradients are available symbolically or can be estimated numerically with a relatively
low computational cost. The most common example of an explicit constraint is the box
constraint, in which each coordinate of the design variable $\xx$ is constrained in a
closed interval. Explicit constraints often appear in engineering optimization as
prerequisites for executing the simulation to compute the objective function. Taking
into account the abovementioned situations, in this study, we assume that
\begin{itemize}\setlength\itemsep{0em}
\item the computational time for $g_j(\xx)$ and its (numerical) gradient $\nabla
g_j(\xx)$ are negligible compared to that for $f(\xx)$; and
\item the objective function $f(\xx)$ is not necessarily defined for an infeasible
solution $\xx$, which violates some $g_j(\xx)$.
\end{itemize}

The covariance matrix adaptation evolution strategy (CMA-ES)
\citep{Hansen2001ec,Hansen2003ec,hansen2016tutorial} is employed in this study as the
baseline optimization method for solving the problem \eqref{eq:orig-prob}. The CMA-ES
is known as one of the state-of-the-art zeroth-order optimization algorithms for
unconstrained black-box continuous optimization. In particular, it has been
demonstrated as efficient for difficult objective functions such as non-convex,
ill-conditioned, and non-separable functions. Since these difficulties of objective
functions naturally appear in engineering optimization, whether or not constraints
exist, it is desirable for search algorithms for constrained optimization to operate
efficiently under these difficulties. Although the CMA-ES is designed for
unconstrained optimization, it can be applied to constrained optimization with the aid
of constraint-handling techniques (CHTs).

One key characteristic of the CMA-ES is its \emph{invariance} to several
transformations of the objective function and search space
\citep{Hansen2011impacts}. The invariance properties induce equivalent problem
classes. All instances (problems) in an equivalent class are regarded as
equivalent under the corresponding transformation of the initial search
distribution. Owing to the invariance properties of the CMA-ES, if its initial
state is transformed properly, it is empirically observed that the CMA-ES
minimizes ill-conditioned and non-separable quadratic functions as efficiently as
well-conditioned and separable spherical functions. This is key to the success of
the CMA-ES in real-world problems because it is designed to adapt the distribution
to such a proper state. Although the invariance properties themselves do not imply
algorithm efficacy, they are useful for generalizing observations. That is, they
are essential for assessing the performance of algorithms empirically. The
invariance further contributes to the quasi-parameter-free feature of the CMA-ES,
which is an important characteristic of the CMA-ES that has attracted attention
from practitioners. As opposed to many other evolutionary approaches to continuous
black-box optimization, in which the hyper-parameters are required to be tuned
depending on the problem characteristics to enable efficient performance
\citep{Karafotias2015}, default values are prepared for all hyper-parameters of
the CMA-ES, depending only on the search space dimension $n$.
As many advantages of the
CMA-ES for unconstrained optimization originate from its invariance properties, \nnew{we
  hypothesize that}{} a
CHT for explicitly constrained optimization is desirable to preserve the
invariance as much as possible to use the CMA-ES for constrained optimization.

However, the CHTs for explicit constraints employed in variants of the ES are not
designed to preserve the invariance of the baseline search algorithms, as we
discuss briefly in \Cref{sec:guideline}. When the CMA-ES is applied to an
explicitly constrained black-box optimization problem with these CHTs, it loses
the invariance properties exhibited by the CMA-ES.
\del{As many advantages of the
CMA-ES for unconstrained optimization originate from its invariance properties, a
CHT for explicitly constrained optimization is desirable to preserve the
invariance as much as possible to use the CMA-ES for constrained optimization.}{}%
\del{The search performance may be
  deteriorated by transformation of the search space or the objective function,
  for which the CMA-ES is invariant.
  CHTs exist that are applicable to the CMA-ES and preserve its invariance
  properties, such as the death penalty or resampling techniques. 
}{}%
\new{
  Certain CHTs, such as death penalty or resampling techniques, are applicable to
  the CMA-ES and preserve its invariance properties.   
}%
However, as these approaches do not exploit the fact that the constraint
violations are cheap to evaluate and their gradient information is available, they
are often inefficient, as observed in \Cref{sec:exp1}.

We propose a novel CHT for the CMA-ES in solving the explicitly
constrained optimization problem \eqref{eq:orig-prob}, named adaptive ranking-based
constraint handling (ARCH).  ARCH is designed to include the following two types of
invariance properties: invariance to the element-wise increasing transformation of the
objective and constraint functions, and invariance to the affine transformation of the
search space.  ARCH replaces the evaluation step in the sampling--evaluation--update
cycle of the CMA-ES, as follows: A candidate solution generated by the CMA-ES is first
repaired on a boundary of the feasible domain.  The objective function value is
evaluated at the repaired point.  A penalty for the repair operation is computed by
the Mahalanobis distance between the original and repaired solutions under the
covariance matrix of the current search distribution.  The candidate solutions are
ranked based on the adaptive weighted sum of the rankings of the objective function
values and rankings of the penalty values.  An adaptive penalty coefficient is
introduced to control the balance between the rankings, and is adapted for the search
distribution so as not to move away from the feasible domain in terms of the
Mahalanobis distance, while allowing infeasible but near boundary solutions to exhibit
high rankings.

The contributions of this study are summarized as follows. Firstly, we present
ARCH, which can handle explicit and nonlinear constraints.%
\footnote{\new{
    Our implementation of ARCH is available in a GitHub repository
    (\url{https://github.com/naoking158/ARCH}).
  }}%
We prove that ARCH is
invariant to any element-wise increasing transformation of $f$ and $g_j$, and to
any affine transformation of the search space coordinates. To the best of the
authors' knowledge, this is the first approach for explicit constraints that is
invariant to these transformations. Secondly, we empirically evaluate the
effectiveness of the proposed approach from an invariance perspective. We develop
test problems to demonstrate the effects of the invariance to affine
transformation. In test cases, we empirically observe that ARCH is invariant to
affine transformations, and illustrate that ARCH is even more effective for box
constrained optimization problems than an existing CHT specialized for the box
constraint. Thirdly, we compare ARCH with existing CHTs for non-explicit
constraints on problems in which the objective function is defined for the
infeasible domain. We use the CEC 2006 constrained optimization benchmark suite to
demonstrate that ARCH overwhelmingly outperforms other CHTs for non-explicit
constraints. This indicates that the explicit constraints should be treated as
explicit constraints even if the objective function is defined on an infeasible
domain. Compared to the previous publication in \citet{sakamoto2019gecco}, we $i$)
improve the algorithm to prevent the search distribution from being unnecessarily
biased toward the boundary if the population size is larger than the default,
$i\!i$) prove the invariance properties of ARCH, and $i\!i\!i$) compare ARCH with
existing approaches on the CEC2006 testbed.

The remainder of this paper is organized as follows. We summarize the mathematical
notations applied throughout the paper below. \Cref{sec:related} introduces the
baseline optimization algorithm, namely the CMA-ES, and related CHTs.  We discuss
the invariance properties for constrained optimization problems in
\Cref{sec:guideline}. Our proposed CHT, ARCH, is described in \Cref{sec:arch}. The
invariant properties of ARCH are demonstrated in \Cref{sec:invariance}. In
\Cref{sec:exp1}, we empirically demonstrate how the invariance properties operate
in practice, by means of numerical experiments on linearly constrained quadratic
problems, to observe the efficacy of ARCH. We present our comparison of ARCH with
other CHTs using the CEC 2006 constrained optimization testbed in \Cref{sec:exp2}.
The paper is concluded in \Cref{sec:conclusion}.

\paragraph{Notations}

In the following, $\R$ is the set of real numbers and $\Rp$ is the set of strictly
positive real numbers.  Let $\xx \in \R^n$ be an $n$-dimensional column vector, where
$\xx^\T$ is its transpose, $\| \xx \|$ denotes the Euclidean norm of $\xx$, and
$[\xx]_i$ denotes the $i$th coordinate of $\xx$. Note that the $i$th coordinate of the
$k$th vector $\xx_k$ is denoted by $[\xx_k]_i$. The ($i$, $j$)th element of a matrix
$\bm{A}$ is also denoted by $[\bm{A}]_{i,j}$. The identity matrix is denoted by
$\I$. The indicator function $\1{\texttt{condition}}$ returns 1 if \texttt{condition}
is true, and 0 otherwise. The sign function $\sgn(a)$ returns 1 if $a > 0$, $-1$ if $a
< 0$, and 0 otherwise.  The integer interval between and including $a$ and $b$ is
denoted by $\llbracket a, b \rrbracket$.

\section{CMA-ES and Related CHTs}
\label{sec:related}


In this section, we introduce the CMA-ES, which is our baseline algorithm for
unconstrained continuous optimization, followed by an overview of the existing CHTs
for the CMA-ES.


\begin{table}[t]
  \begin{center}
    \caption{Default parameter setting for CMA-ES.}
    \label{tab:defaultparameter}
    \begin{tabular}{c}
      \toprule
      $\lambda = 4 + \lfloor 3 \ln(n) \rfloor, \quad
      \mu = \lfloor \lambda / 2 \rfloor$ \\
      \multirow{2}{*}{$\ww_i = \frac{\ln (\frac{\lambda+1}{2}) - \ln(i)}{
          \sum^{\mu}_{k=1} \left( \ln (\frac{\lambda+1}{2}) - \ln(k) \right) } , \quad
      \cs = \frac{ \muw + 2 }{ n + \muw + 5 }, \quad
      \cc = \frac{ 4 + \muw / n }{ n + 4 + 2\muw / n }$}\\
      \\
      \multirow{2}{*}{$\cone = \frac{2}{( n + 1.3 )^2 + \muw }, \quad
      \cmu = \min \left(1-\cone, \frac{2(\muw - 2 + 1 / \muw)}{(n+2)^{2} + \muw } \right)$}\\
      \\
      \multirow{2}{*}{$\ds = 1 + \cs + 2 \times \max \left(0, \sqrt{\frac{\muw - 1}{n + 1}} - 1 \right)$}\\
      \\
      \bottomrule
    \end{tabular}
  \end{center}
\end{table}

\subsection{CMA-ES}
\label{subsec:cmaes}

The CMA-ES~\citep{Hansen2003ec,hansen2016tutorial} is a stochastic multi-point search
algorithm for black-box continuous minimization of $f: \R^n \to \R$. The CMA-ES
samples $\lambda$ candidate solutions $\xx_k$ for $k \in \{1, \dots, \lambda\}$ from
the multivariate normal distribution $\dist$, where $\mm \in \R^n$ is the mean vector,
$\sigma \in \Rp$ is the step size, and $\boldsymbol{C} \in \R^{n \times n}$ is the
covariance matrix. These distribution parameters are updated using the candidate
solutions and their ranking information.

\step{0}.  Initialize $\boldsymbol{m}^{(0)}, \sigma^{(0)}, \boldsymbol{C}^{(0)}$
according to the initial problem search domain, and initialize two evolution paths
$\boldsymbol{p}_c^{(0)} = \boldsymbol{p}_\sigma^{(0)} = \boldsymbol{0}$ and their
correction factors $\gs^\ite{0} = \gc^\ite{0} = 0$. All
parameters that appear in
the following are set to the default values listed in \Cref{tab:defaultparameter}. The
meanings of these parameters are described in \citet{hansen2016tutorial}. These are
designed based on theoretical ES research (e.g., see \citet{AAH2020}) and extensive
experiments. The CMA-ES repeats the following steps at each iteration, $t = 0, 1,
\cdots$, until a termination criterion is satisfied.

\step{1}.  Draw $\lambda$ samples $\zz_k$ for $k \in \{1, \dots, \lambda\}$
independently from $\mathcal{N}(\boldsymbol{0}, \I)$. Compute $\yy_k =
\sqrt{\cov^\ite{t}} \zz_k$ and $\xx_k = \mm^\ite{t} + \sigma^\ite{t} \yy_k$. Then,
$\xx_k$ ($k = 1, \dots, \lambda$) are the candidate solutions that are independently
$\mathcal{N}(\mm^\ite{t}, (\sigma^\ite{t})^2 \cov^\ite{t})$ distributed. Here,
$\sqrt{\cov^\ite{t}}$ is the symmetric matrix satisfying $\cov^\ite{t} =
\left(\sqrt{\cov^\ite{t}} \right)^2$.

\step{2}.  Evaluate the candidate solutions $\xx_k$, for $k \in \{1, \dots,
\lambda\}$, on the loss function $L$, and sort them in ascending order. In an
unconstrained optimization scenario, usually, $L = f$. Let the $i$th best candidate
solution be denoted by $\boldsymbol{x}_{i:\lambda}$. In the same manner, we denote the
corresponding steps and normalized steps as $\boldsymbol{y}_{i:\lambda}$ and
$\boldsymbol{z}_{i:\lambda}$, respectively.

\step{3}.  Compute the weighted sum of the $\mu$ best steps of the candidate solutions
$\dy = \sum^\mu_{i=1} {\rm w}_i \boldsymbol{y}_{i:\lambda}$ and update the mean vector
$\boldsymbol{m}^{(t)}$, as follows:
\begin{align*} \boldsymbol{m}^{(t+1)} = \boldsymbol{m}^{(t)} + \sigma^{(t)} \dy
\enspace,
\end{align*}%
where ${\rm w}_i$ is the recombination weight for the $i$th best
candidate, which satisfies ${\rm w}_1 \geq {\rm w}_2 \geq \dots \geq {\rm w}_\mu > 0$
and $\sum^\mu_{i = 1} {\rm w}_i = 1$.

\step{4}.
Update the evolution paths according to
\begin{align*}
  \ps^\ite{t+1} &= (1 - \cs) \ps^\ite{t}
                  + \sqrt{\cs ( 2 - \cs ) \mu_\ww} \Big(\sqrt{\cov^\ite{t}}\Big)^{-1} \dy\enspace,\\
  \pc^\ite{t+1} &= (1-\cc) \pc^\ite{t} + \hsig^\ite{t+1} \sqrt{\cc ( 2 - \cc ) \mu_\ww} \dy \enspace,
\end{align*}%
where $\cs$ and $\cc$ are the cumulation factors for the evolution paths, $\mu_\ww = 1
/ \sum^\mu_{i=1} \ww_i^2$,
\begin{align*}
  \hsig^\ite{t+1} =
  \begin{cases}
    1 & \text{if $\norm{\ps^\ite{t+1}} < \big(1.4 + \frac{2}{n+1}\big) \big(\gs^\ite{t+1}\big)^{\frac{1}{2}} \chi$} \\
    0 & \text{otherwise},
  \end{cases}
\end{align*}%
and $\chi = \mathbb{E}[ \| \mathcal{N}(\boldsymbol{0}, \I) \| ] \approx n^{\frac12}
\big( 1 - \frac{1}{4n} + \frac{1}{21n^2} \big)$ is the expectation of the norm of the
$n$ variate standard normal distribution.  The Heaviside function $\hsig^\ite{t+1}$
stalls the update of $\pc^\ite{t+1}$ if $\norm{\ps^\ite{t+1}}$ is large. The
correction factors for the evolution paths\footnote{
  Note that we introduce $\gs$ and $\gc$ that does not appear in a standard
formulation in order to treat the initialization effect of the evolution paths and
write it short. See \citet{akimoto2019ddcma} for more detail.
} are updated as follows:
\begin{align*}
  \gs^\ite{t+1} &= (1 - \cs)^2 \gs^\ite{t} + \cs ( 2 - \cs) \enspace , \\
  \gc^\ite{t+1} &= (1 - \cc)^2 \gc^\ite{t} + \hsig^\ite{t+1} \cc (2 - \cc) \enspace.
\end{align*}%

\step{5}.
Update the step size and covariance matrix as follows:
\begin{align*}
  \sigma^\ite{t+1} &= \sigma^\ite{t} \exp \left( \frac{ \cs }{ \ds }
                     \left( \frac{\norm{\ps^\ite{t+1}}}{\chi} - \big(\gs^\ite{t+1}\big)^\frac12 \right) \right) \enspace, \\
  \cov^\ite{t+1}  &= \cov^\ite{t}
                    + \cone \left( \pc^\ite{t+1} (\pc^\ite{t+1})^\T - \gc^\ite{t+1} \cov^\ite{t} \right)
                    + \cmu \sum^\mu_{i=1} \ww_i \left( \yy_{i:\lambda} (\yy_{i:\lambda})^\T - \cov^\ite{t} \right) \enspace,
\end{align*}%
where $\ds$ is the damping parameter for the step size adaptation, while $\cone$ and
$\cmu$ are the learning rate for the rank-one and rank-$\mu$ updates of the covariance
matrix, respectively.

\subsection{CHTs for ESs}
\label{ssec:cht}

We briefly review the CHTs employed in variants of ESs.

\subsubsection{Resampling and Death Penalty}


The resampling technique is the simplest CHT.  Candidate solutions are resampled
repeatedly until $\lambda$ feasible candidate solutions are generated.  To guarantee
that the resampling stops within a finite time, the maximum number of sampling in one
iteration is set to a finite number. If the number of feasible candidate solutions is
less than $\lambda$, the current population is filled with infeasible solutions and
the worst loss value is assigned to them (for example, $+\infty$).  When the maximum
sampling number is set to $\lambda$, the resampling technique is known as the death
penalty.

The resampling and death penalty methods are easy to implement and are applicable to
any constraint type. However, they are not appropriate if the optimum is located on
the boundary of the feasible domain, as candidate solutions are biased in the feasible
domain and the search distribution tends to approach the boundary slowly, as we
observe in \Cref{sec:exp1}.  Moreover, the search algorithm cannot conduct a
meaningful ranking of the candidate solutions if the probability of sampling feasible
solutions is rather low and the population is filled with infeasible solutions.  If
this is the case, the parameter update will result in random fluctuation.

\subsubsection{Penalty Function Methods}
\label{sssec:penalty}

CHTs based on the penalization of infeasible candidate solutions are the most
extensively used CHTs for real-world engineering optimization problems. The main
concept of penalty function methods is transforming a constrained optimization problem
into an unconstrained optimization problem by defining the penalized loss function:
\begin{equation} L(\xx) = f(\xx) + p(\xx) \enspace,
\end{equation} where $p$ is a penalty function. If the objective function value is not
well defined for infeasible solutions, a repair operation needs to be applied, and the
first term is replaced with the objective function value evaluated at the repaired
solution. The penalty function is manually designed in many engineering optimization
problems, and a typical choice is the weighted sum of the constraint violations
$\max(g_j(\xx), 0)$ or its monotone transformation, such as square.

The adaptive penalty box constraint handling (AP-BCH) method~\citep{Hansen2009tec} is
a penalty function-based box CHT. The loss function is defined as
\begin{align}
  L(\x) = f(\xfeas) + \frac{1}{n} \sum^{n}_{i=1} \gamma_i ([\x]_i - [\xfeas]_i)^2
  \enspace,
\end{align}%
where $\gamma_i$ is an adaptive penalty coefficient and $\xfeas$ is a feasible vector
closest to the infeasible solution $\xx$; that is, $\xfeas = \argmin_{\yy} \norm{\x -
\yy}$.  The feasible solution $\xfeas$ is only used for evaluating the objective
function and the penalty; that is, the repair operator is used in the Darwinian
manner.  This approach does not assume that the objective function is well defined
outside the feasible domain, but it is only applicable to box-constrained problems.

The adaptive augmented Lagrangian constraint handling (AL)
method~\citep{Arnold:2015:TAL:2739480.2754813,atamna2016ppsn} adapts the augmented
Lagrangian
\begin{align*}
  L(\x) = f(\x) + \sum^{m}_{j=1}
  \begin{cases}
    \gamma_j g_j(\x) + \frac{\omega_j}{2} g_j^2(\xx)  & \text{if $\gamma_j + \omega_j g_j(\xx) \leq 0$} \enspace, \\
    - \frac{\gamma_j^2}{2 \omega_j} & \text{otherwise}\enspace,
  \end{cases}
\end{align*}%
where $\gamma_j \in \R$ is a Lagrange factor, $\omega_j \in \R^+$ is a penalty
coefficient, and both are adapted during the optimization process.  AL was initially
proposed for ($1+1$)-ES~\citep{Arnold:2015:TAL:2739480.2754813}, and was extended to
the CMA-ES in a single constraint case~\citep{atamna2016ppsn}, where the median
success rule was applied for the step-size adaptation. The AL is designed for
\emph{implicit} constraints, in which the \emph{implicit} constraint means that the
evaluation is expensive or is computed at the same time as the objective function.  It
requires the objective function to be defined in the infeasible domain.

\subsubsection{Ranking-based Methods}
\label{sssec:ranking}

Ranking-based CHTs aggregate the rankings of the objective function values and
constraint function values to create the final rankings of the candidate solutions,
instead of aggregating the function values.  The stochastic ranking technique
\citep{Runarsson2000tevc} attempts to balance the objective and constraint functions
by sorting the candidate solutions according to the objective function with a
probability $P_f$, and the sum of the constraint function values with a probability $1
- P_f$.  The multiple constraint ranking (MCR) technique~\citep{garcia2017mcr} ranks
the candidate solutions according to the sum of the rankings of the objective values
and rankings of each constraint violation value.  Other techniques included in
ranking-based CHTs such as the lexicographic ordering and the $\eps$-lexicographic
ordering have been proposed in GA and DE, and imported to ES
\citep{Oyman1999,Hellwig2018}.  The advantage of these approaches over penalty
function-based approaches is that they are invariant to strictly increasing
transformation of the objective function and constraint functions, which we later
refer to as \emph{element-wise increasing transformation}. Therefore, practitioners do
not need to tune the balance between the objective function values and constraint
violation values manually.  As with the AL, however, this method requires the
objective function to be defined in the infeasible domain.

\subsubsection{Active Constraint Handling}
\label{sssec:ach}

There are certain CHTs that modify the covariance matrix adaptation mechanism to
shrink the variance actively in the direction of the constraint function gradient so
as to decrease the likelihood of infeasible solutions.  We refer to such methods as
active constraint handling (ACH) in this paper.  Reference~\citet{Arnold:2012:gecco}
proposed the use of the active covariance matrix update in the (1+1)-CMA-ES.  Similar
concepts have been employed in other variants of the CMA-ES, such as the ($\mu,
\lambda$)-CMA-ES in \citet{chocat2015modifiedCMA}, MA-ES in \citet{Spettel2019},
xCMA-ES in \citet{Krause2015}.  As ACH techniques use binary information, whether or
not the constraint is violated, they are invariant to monotone transformations of the
objective and constraint violations. Such methods can be applied to problems in which
the constraints return the outcome that the given solution is either feasible or
not. However, this approach tends to be inefficient compared to other CHTs on
quantifiable constraints, as it does not utilize the amount of constraint violations.

\subsubsection{Other Explicit CHTs}
\label{sssec:others}

The linear constraint covariance matrix self-adaptation ES
(lcCMSA-ES)~\citep{Spettel2018} handles explicit and linear constraints in a variant
of CMA-ES, known as CMSA-ES. It basically samples only feasible solutions, and updates
the distribution parameters by using the feasible solutions. Active-set
ES~\citep{Arnold2016ppsn,Arnold2017gecco} is also designed for \emph{explicit}, but
not necessarily linear, constraints.  This approach is the most relevant one from the
perspective of the assumptions on the constraint problem. It is a (1+1)-ES based
approach that applies a repair operator in the Lamarckian manner; that is, the repair
solution is used as the candidate solution, and not only to compute the loss function
value. Unfortunately, it is not possible to directly extend it to the state-of-the-art
variant of the ES, namely ($\mu$, $\lambda$)-CMA-ES.

\subsection{Formal Classification of CHTs}
\label{ssec:taxonomy}

We summarize the formal classification of the abovementioned CHTs in \Cref{tab:class}.
The taxonomy proposed in \citet{2015arXiv150507881L} clasifies constraints as follows:
\textbf{Q}uantifiable/\textbf{N}onquantifiable,
\textbf{R}elaxable/\textbf{U}nrelaxable, \textbf{S}imulation-based/\textbf{A} priori,
and \textbf{K}nown/\textbf{H}idden.  A \textbf{Q}uantifiable constraint is a
constraint for which the amount of feasibility and/or violation can be quantified,
while a \textbf{N}onquantifiable constraint returns a binary output indicating whether
or not the solution satisfies the constraint.  A \textbf{R}elaxable constraint is a
constraint that does not need to be satisfied to compute the objective function, while
an \textbf{U}nrelaxable constraint is a prerequisite for executing the simulation for
the objective function computation.  An \textbf{A} priori constraint is a constraint
for which the feasibility can be confirmed without running a simulation; that is, this
constraint can be formulated using optimization variables such as $g(\x) =
\sum^{n}_{i=1} [\x]_i \leq 1$, while a \textbf{S}imulation-based constraint is only
computed through a computationally expensive simulation.  A \textbf{K}nown constraint
is a constraint that is explicitly provided in the problem formulation; for example,
$\min f(\x)$ s.t. $g(\x) \leq 0$.  Constrained problems can be expressed by combining
these initial letters as an acronym, such as QRSK.  Refer to
\citet{2015arXiv150507881L} for further descriptions of each type of constraint and
example situations.  Our assumption on the constraints is QUAK in this terminology.

CHTs assuming \textbf{U}nrelaxable constraints can be applied to QUAK constraints. For
example, the resampling technique can be applied to QUAK constraints. However, CHTs
assuming weaker conditions on constraints utilize less information on the constraints
than that available to the optimization approaches. Therefore, we expect that CHTs
assuming QUAK constraints are more efficient for solving QUAK constrained optimization
problems. Only two approaches in \Cref{tab:class}, including the proposed approach,
are designed for QUAK nonlinear constraints. As it is not clear how active-set ES is
extended to a variant of the CMA-ES, the proposed approach is the only approach
applied to the CMA-ES. However, if the constraints are nonlinear but
\textbf{R}elaxable, many of the CHTs listed in \Cref{tab:class} can be applied.

\begin{table}[t]
  \centering
  \caption{Classification of CHTs, where bnds/lc/nlc means that the CHT can handle
bound constraints, linear constraints or nonlinear constraints, respectively.}
  \label{tab:class}
  \small
  \begin{tabular}{l|c|l}
    \hline
    \textbf{CHT (bnds/lc/nlc)}                                          & \textbf{Taxonomy} & \multicolumn{1}{c}{\textbf{Invariance}}    \\
    \hline \hline
    ARCH [proposed CHT, described in \Cref{sec:arch}] (nlc)              & QUAK              & increasing                      / affine   \\
    AP-BCH \citep{Hansen2009tec} (bnds)                                  & QUAK              & {\; \; \; \; \; \; \; $\times$} / $\times$ \\
    lcCMSA-ES \citep{Spettel2018} (lc)                                   & QUAK              & increasing                      / $\times$ \\
    Active-set ES \citep{Arnold2017gecco} (nlc)                          & QUAK              & increasing                      / $\times$ \\
    AL \citep{atamna2016ppsn} (nlc)                                  & QRSK              & {\; \; \; \; \; \; \; $\times$} / affine   \\
    Stochastic ranking \citep{Runarsson2000tevc} (nlc)                   & QRSK              & {\; \; \; \; \; \; \; $\times$} / affine   \\
    MCR \citep{garcia2017mcr} (nlc)                                      & QRSK              & increasing                      / affine   \\
    (1+1)-CMA-ES with ACH \citep{Arnold:2012:gecco} (nlc)              & NUSK              & increasing                      / affine   \\
    ($\mu, \lambda$)-CMA-ES with ACH \citep{chocat2015modifiedCMA} (nlc) & NRSK              & increasing                      / affine   \\
    xCMA-ES with ACH \citep{Krause2015} (nlc)                            & NUSK              & increasing                      / affine   \\
    MA-ES with ACH \citep{Spettel2019} (nlc)                             & NUSK              & increasing                      / affine   \\
    Resampling technique (nlc)                                          & NUSH              & increasing                      / affine   \\
    \hline
  \end{tabular}
\end{table}

\section{Desired Invariance Properties for Constrained Optimization}
\label{sec:guideline}


We describe two invariance properties that are desirable for a CHT that is designed
for variants of the CMA-ES\ddel{ to preserve their effectiveness}{}.

\subsection{Element-wise Increasing Transformation of Functions}
\label{ssec:increasing}

An strictly increasing transformation $h: \R \to \R$ is a function satisfying $h(t) <
h(s)$ if $t < s$.  Invariance to an increasing transformation of the objective
function in an unconstrained optimization scenario refers to the property whereby the
algorithm does not change the behavior (which is possibly characterized by the
sequence of the solutions generated by the algorithm) when solving $f$ and its
composite $h \circ f$.  Algorithms that are invariant to any increasing transformation
can solve, for example, a non-convex discontinuous function $h \circ f$ as easily as a
convex continuous functions $f$. The importance of this invariance property is
extensively recognized in engineering optimization: if the search algorithm is not
invariant, the objective function needs to be tuned for the search algorithm to
perform effectively, which is time consuming. Numerous evolutionary algorithms,
including the CMA-ES, are invariant to any increasing transformation of the objective
function, because they use only the objective function value rankings of the candidate
solutions.

In constrained optimization, we consider invariance to an element-wise increasing
transformation $H = (h_0, \dots, h_m): \R^{m+1} \to \R^{m+1}$ of the objective and
constraint functions $F = (f, g_1, \dots, g_m)$, where $h_j: \R \to \R$
($j=0,\dots,m$) is an increasing transformation and $h_j$ for $j = 1,\dots, m$
satisfies $h_j(0) = 0$. Invariance to an element-wise increasing transformation of the
objective and constraint functions refers to the property whereby the algorithm does
not change the behavior when solving a constrained problem $F = (f, g_1, \dots, g_m) =
(h_0 \circ f, h_1 \circ g_1, \dots, h_m \circ g_m) = H \circ F$.  The original
constrained optimization problem $F$ and its transformation $H\circ F$ models the same
optimization problem in the sense that they define the same feasible domain $X$ and
the same total order on $X$ regarding the objective function value \footnote{ This
transformation is applied after the transformation from the equality constraint to the
inequality one described in the introduction is performed.  } .

In real-world applications, the ranges of the objective function values and constraint
violations are often quite different. Algorithms without this invariance will suffer
from this difference and place implicit priority on an objective or certain
constraints depending on their values. Therefore, practitioners may determine a
reasonable transformation $H$. As described above for unconstrained optimization
cases, this can be time consuming, and requires domain knowledge of the problem and
deep insight into the optimization algorithm.

Although this invariance property is a straightforward extension of invariance to the
increasing transformation of the objective function and it is seemingly important, it
is not exhibited by the frequently used penalty function-based techniques that take
the sum of the objective and constraint function values as loss values. For example,
it is clear that the AL does not exhibit this invariance: it is not invariant to
increasing the transformation of $f$ and $g_j$.  Although the AP-BCH is invariant to
any increasing transformation of $g_j$, it is not invariant to an increasing
transformation of $f$ in general, as the quadratic penalty term is added to the
objective function value directly.  It has been demonstrated in
\citet{Sakamoto2017geccoposter} that the AP-BCH deteriorates when the objective
function is, for example, an exponential function, where the objective function value
is more sensitive than the quadratic penalty term.

\subsection{Affine Transformation of Search Space Coordinates}\label{ssec:affine}
\providecommand{\pp}{\bm{p}} \providecommand{\qq}{\bm{q}}

To formulate the invariance properties for constrained optimization problems, we
consider the fact that our minimization problem is defined on an $n$-dimensional inner
product space $(V, \inner{\cdot}{\cdot})$ on the real field $\R$:
\begin{equation}\label{eq:vectorspace} \argmin_{\pp \in V} f^V(\pp) \ \text{s.t.} \
g^{V}_{j}(\pp) \leq 0,\ \forall j = 1,\dots, m \enspace,
\end{equation} where $f^V: V \to \R$ and $g^V_j: V \to \R$ are the objective and
constraint functions, respectively. The constrained problem \eqref{eq:orig-prob} is
considered a realization of \eqref{eq:vectorspace} under an orthonormal basis
$\{\bm{e}_i \in V\}_{i=1}^{n}$ and a bias vector $\bm{e}_0 \in V$, where
$\inner{\bm{e}_i}{\bm{e}_i} = 1$ for $i = 1,\dots,n$ and $\inner{\bm{e}_i}{\bm{e}_j} =
0$ for any $j \neq i$ ($i,\ j \geq 1$), and $\xx \in \R^n$ corresponds to $\pp =
\bm{e}_0 + \sum_{i=1}^{n}[\xx]_i \bm{e}_i \in V$. Let $\pp \mapsto \xx$ be denoted by
$\psi_e: V \to \R^n$. If the optimization algorithm is defined on the inner product
space, its behavior (which is possibly characterized by the sequence of the solutions
generated by the algorithm) is identical on any basis $\{\bm{v}_i \in V\}_{i=1}^{n}$
with a bias vector $\bm{v}_0 \in V$, where the bases are not necessarily orthonormal
to one another. Let $\pp = \bm{v}_0 + \sum_{i=1}^{n}[\tilde\xx]_i \bm{v}_i \mapsto
\tilde\xx$ be denoted by $\psi_v: V \to \R^n$. The map $A:\psi_e \circ \psi_v^{-1}$
from a coordinate system $\psi_v$ to a coordinate system $\psi_e$ is an affine
transformation. The invariance to an affine transformation of the search space
coordinates refers to the property whereby the algorithm behaves the same on the
original coordinate system and its affine transformed coordinate system.

The objective function $f^V(\pp) = \frac12 \inner{\pp}{\pp}$ is expressed as
$f^{e}(\xx) = f^V(\psi_e^{-1}(\xx)) = \sum_{i=1}^{n}[\xx -
\psi_e(\bm{e}_0)]_{i}^2$ in the system $\psi_e$, while it is expressed as
$f^{v}(\tilde\xx) = f^e(A(\tilde\xx)) = \sum_{i=1}^{n}[A \tilde\xx -
\psi_e(\bm{v}_0)]_{i}^2$ in the system $\psi_v$. The former is the sphere function
(well conditioned and separable), while the latter is a convex quadratic function
that is ill conditioned if $A$ also is, and is non-separable if $A$ is not
diagonal. If the algorithm is invariant to any affine transformation of the search
space coordinates, the algorithm solves the ill-conditioned and non-separable
function $f^v$ as efficiently as it solves the well-conditioned and separable
$f^e$ under the corresponding transformation of the initial search distribution.
Not all variants of the CMA-ES, including that presented in~\Cref{subsec:cmaes},
are (proven to be) invariant to any affine transformation of the search space
coordinate\footnote{The invariance of CMA-ES can be proven by using $\norm{\left(
\sqrt{\cov^\ite{t}} \right)^{-1} \pc}$ instead of $\norm{\ps}$ of the step size
update formula in \Cref{subsec:cmaes} (note the meaning changes because what is
accumulated is not z). However, the step size formula using $\ps$ is generally
employed, and no adverse effect of this difference has been empirically
observed. Therefore, we use $\ps$ as well in this paper.}, although we empirically
observe statistically invariant behaviors under arbitrary affine transformations
(see \cref{fig:medfig}). Numerous other evolutionary computation approaches are
not invariant to these transformations, and we empirically observe rather
different performances depending on the affine transformation properties.

In real-world applications, the change in the coordinate system corresponds to the
change in the features describing the object to be optimized, or the change in the
unit in each feature.\ddel{ It is often difficult to determine reasonable feature
vectors and their units in advance of the optimization process. The CMA-ES
  is designed to adapt its state to the coordinate system; afterward, it
\del{%
  can efficiently search
}{}%
{%
  is expected to behave identically on different coordinate systems
}%
because of the invariant to affine transformations.  Therefore, together with the
adaptation ability, invariance to the affine transformation is key to the success
of the CMA-ES\del{, and its importance is well recognized}{}.
}{}
In a constrained optimization scenario, the constraint function $g_j$ and the
objective function $f$ are transformed by the same transformation $A$, resulting in
$g_j \circ A$ and $f \circ A$ in the transformed coordinate system, respectively. The
linear constraints are again linear in the transformed coordinate system. However, a
box constraint will no longer be a box constraint, but rather a set of linear
constraints (forming an $n$-parallelotope shape).

Suppose that the underlying unconstrained optimization algorithm is invariant to any
affine transformation of the search space coordinates. As summarized in
\Cref{tab:class}, CHTs that only touch the loss function values, such as the
resampling technique, ACH, and AL, do not disturb the invariance property of the
underlying algorithm, resulting in invariance to the affine transformation of the
search space under the constraints (see \citet{atamna2020} for proof of the affine
invariance of AL). However, CHTs for unrelaxable constraints that require a repair
operator often lose the invariance property. For example, the AP-BCH does not exhibit
this invariance property, as the repair operator used in this approach exploits the
fact that the feasible domain is composed of an interval in each coordinate. Repair
operators that use the inner product or the distance in Euclidean space are generally
affected by the affine transformation of the coordinate system.

\section{ARCH}
\label{sec:arch}


We propose an explicit constraint-handling method based on adaptive ranking, known as
ARCH.  ARCH virtually transforms the constrained optimization problem into an
unconstrained problem by defining an adaptive loss function $L$. The loss $L$ is
defined by the weighted sum of the rankings of the objective and constraint
violations, denoted as $L = \RT$ below.  The proposed algorithm is designed to exhibit
the invariance properties listed in \Cref{sec:guideline}, and it does not require the
objective function to be well defined in the infeasible domain.

\subsection{Repair Operator}
\label{ssec:repair}

To make ARCH applicable to a problem in which the objective function values are not
defined in the infeasible domain, we employ a repair operator in the Darwinian
manner. Given a candidate solution $\xx$, ARCH determines a repaired solution
$\tilde{\xx}$.  The repaired solution $\tilde{\xx}$ is only used for the objective
function value computation.

The repair operator $\mathtt{Repair}(\cdot)$ is defined as follows: Let $\mathcal{S} =
\set{\xx \in \R^n \mid g_j(\xx) \leq 0, \ \forall j \in \llbracket 1, m \rrbracket}$
be the feasible domain.  Given a solution $\xx$, let $\mathcal{J}(\xx) = \set{j \in
\llbracket 1, m \rrbracket \mid g_j(\xx) > 0}$ be the set of indices of the violated
constraints.  Let $\mathcal{A}(\xx) = \set{\yy \in \R^n \mid g_j(\yy) = 0, \ \forall j
  \in \J(\xx)}$ be the intersection of the violated constraint boundaries. We
introduce the repair operator defined as follows:
\begin{empheq}[
  left={\tilde{\xx} = \mathtt{Repair}(\xx) =
    \argmin_{\yy} \ \norm{\xx - \yy}^2_{\Cinv} \quad
    \text{s.t.\ } \ \yy \in \empheqlbrace},
  right={\enspace ,}]{align}
  &\mathcal{A}(\xx) \cap \mathcal{S} & \quad \text{if $\mathcal{A}(\xx) \cap \mathcal{S} \not= \emptyset$}  \label{eq:intersection}
  \\
  &\mathcal{S} & \text{otherwise} \qquad \ \,  \label{eq:nearest}
\end{empheq}%
\begin{wrapfigure}{r}[0pt]{0.4\hsize} \centering
\includegraphics[width=\hsize]{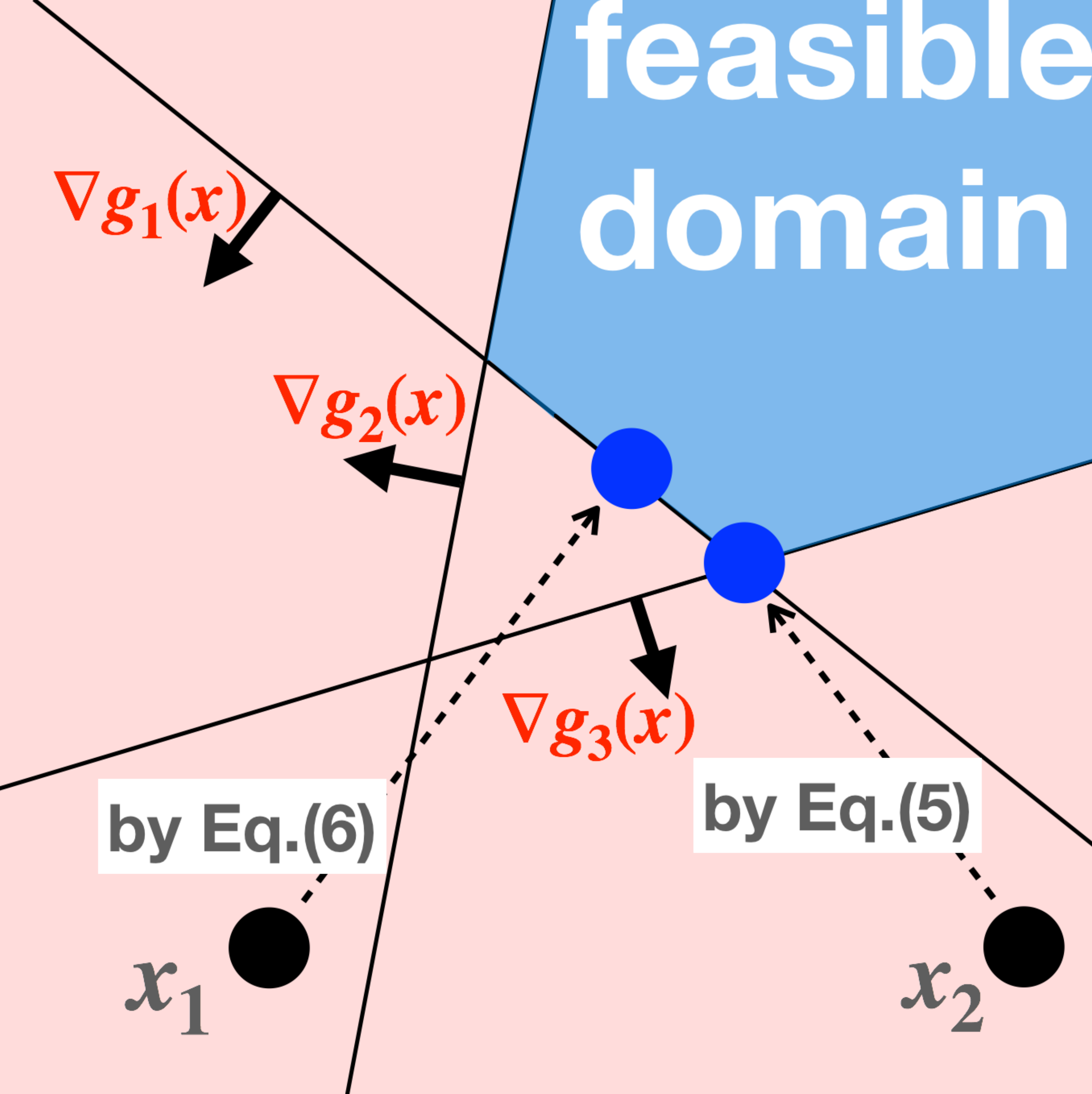}
  \caption{\label{fig:repair_op}For example, $\mathcal{J}(\xx_{1}) = \set{1, 2, 3}$
and $\mathcal{A}(\xx_{1}) = \emptyset$, then $\xx_{1}$ is repaired by using
\cref{eq:nearest}.}
\end{wrapfigure} where $\norm{\xx - \yy}^2_{\Cinv} = (\xx - \yy)^\T \Cinv (\xx - \yy)$
is the Mahalanobis distance between $\xx$ and $\yy$ under the inverse matrix $\Cinv =
(\sigma^2 \CC)^{-1}$.  This is a constrained minimization of a quadratic function,
which is solved by standard numerical optimization routines. If the constraints are
all linear, the problem is reduced to a quadratic programming problem.
\Cref{eq:nearest} is the nearest feasible solution to $\xx$, while
\cref{eq:intersection} is the feasible solution nearest to $\xx$ under the constraint
that the active constraints at $\xx$ remain active at $\tilde{\xx}$.  The reason that
\cref{eq:intersection} is preferred over \cref{eq:nearest} is explained in~\Cref{app:repair}.

The condition that $\mathcal{A}(\xx) \cap \mathcal{S} \neq \emptyset$ is indeed
simplified to $\mathcal{A}(\xx) \neq \emptyset$ when the constraints are all linear
and not redundant.  Here, a redundant constraint $g_j$ is defined as a constraint such
that the boundary of $g_j$ is never in contact with the feasible domain; that is,
$\set{\xx \in \R^n \mid g_j(\xx) = 0} \notin \mathcal{S}$.  Note that determining
whether $\mathcal{A}(\xx) \neq \emptyset$ is as easy as checking whether a system of
linear equations has a solution. In explicit linear constraint situations, such a
redundant constraint can be removed in advance.

In this approach, the Mahalanobis distance is employed rather than the Euclidean
distance $\norm{\xx - \yy}^2 = (\xx - \yy)^\T (\xx - \yy)$, in order to achieve affine
invariance of the search space coordinate system. This point is discussed formally in
\Cref{sec:invariance}.

\subsection{Total Ranking}
\label{ssec:totalranking}

The total ranking $\RT(\xx_k)$ for $k \in \set{1, \dots, \lambda}$ at each iteration
$t \in \set{0, 1, \dots}$ is a weighted sum of the rankings of the candidate solutions
based on the objective function values, $\Rf(\xx_k)$, and based on the Mahalanobis
distance to the feasible domain, $\Rg(\xx_k)$, namely:
\begin{align} \RT(\xx_k) = \Rf(\xx_k) + \alpha^\ite{t+1} \Rg(\xx_k)
  \label{eq:rt} \enspace ,
\end{align}%
where $\alpha^\ite{t+1}$ is known as the ranking coefficient that
controls the balance between the objective and constraints.  The rankings $\Rf(\xx_k)$
and $\Rg(\xx_k)$ are defined as follows:

The $f$-ranking, $\Rf(\xx_k)$, is the number of better candidate solutions in terms of
$f$, plus the number of tie candidate solutions divided by 2, namely
\begin{align}
  \Rf(\xx_k) = \sum^{\lambda}_{l=1}\1{f(\tilde{\xx}_l) < f(\tilde{\xx}_k)}
  + \frac{1}{2} \sum^{\lambda}_{l=1} \1{f(\tilde{\xx}_l) = f(\tilde{\xx}_k)}
  \enspace,
  \label{eq:rf}
\end{align}%
where $\tilde{\xx}_{i} = \mathtt{Repair}(\xx_{i})$. Note that the sum of $\Rf(\xx_k)$
for $k \in \set{1, \dots, \lambda}$ is $\lambda^2 / 2$. Moreover, in an unconstrained
optimization scenario, the probability of sampling tie solutions is often zero, while
the second term above may be nonzero with a nonzero probability in our situation owing
to the repair operation.

The $g$-ranking, $\Rg(\xx_k)$, is analogously defined by simply replacing $f \circ
\mathtt{Repair}$ with $\gsig$:
\begin{align}
  \Rg(\xx_k) = \sum^{\lambda}_{l=1} \1{\gsig(\xx_l) < \gsig(\xx_k)} + \frac{1}{2}
  \sum^{\lambda}_{l=1} \1{\gsig(\xx_l) = \gsig(\xx_k)} \enspace,
  \label{eq:rg}
\end{align}%
where $\gsig(\xx) = \norm{\xx - \tilde{\xx}}^2_\Cinv$ and $\Sigma = (\sigma^\ite{t})^2
\cov^\ite{t}$.

\subsection{Adaptation of Ranking Coefficient}
\label{ssec:adaptation}

The ranking coefficient $\alpha$ controls the balance between $\Rf$ and $\Rg$.  If
$\alpha$ is too large, the search distribution is biased toward the feasible domain.
If $\alpha$ is too small, the search distribution is biased toward the infeasible
domain.  Therefore, the adaptation of $\alpha$ significantly influences the search
performance.

The adaptation of $\alpha$ is based on a theoretical study of the weighted
recombination ES on a spherical function.  It was reported in \citet{AAH2020} that, in
the optimal situation in terms of quality gain~\citep{beyer2001}
\begin{align}
  \frac{\norm{\mm_\mathrm{sph} - \xx^*_\mathrm{sph}}}{n \cdot \sigma}
  = \frac{1}{\snf}
  \approx \frac{1}{\hat{\sigma}},
  \enspace
  \hat{\sigma} = \frac{c \cdot n \cdot \mueff}{n - 1 + c^2 \cdot \mueff}
  \enspace,
  \label{eq:snf}
\end{align}%
where $\mm_\mathrm{sph}$ and $\xx^*_\mathrm{sph}$ are the mean vector and optimum on a
spherical function, respectively; $\snf$ is the optimal normalized step size, which is
approximated by $\hat{\sigma}$; $c = - \sum^{\mu}_{i = 1} \ww_i \E[\NN_{i:\lambda}]$
is the weighted average of the expected value of the normal order statistics from
$\lambda$ samples and is usually in $\mathcal{O}(1)$; and $\mueff = (\sum^{\mu}_{i=1}
\ww_i^2)^{-1}$ is the variance effective selection mass.

Ideally, the CMA-ES with the proposed constraint handling should treat a constrained
sphere function as though it is an unconstrained sphere function.  That is, we wish
\eq{eq:snf} to hold even for a constrained sphere problem.

Assuming that the optimum of a constrained $n$-dimensional sphere problem having $n$
constraints is located on the $n$ boundaries, we estimate the left-hand side of
\eq{eq:snf} using the Mahalanobis distance between $\mm$ violating $n$ constraints and
$\tilde{\mm} = \mathtt{Repair}(\mm)$:
\begin{align}
  \frac{\norm{\mm_\mathrm{sph} - \xx^*_\mathrm{sph}}^2}{\sigma^2}
  \approx \norm{\mm - \tilde{\mm}}^2_\Cinv
  \enspace.
  \label{eq:approx_mdist}
\end{align}%
We define $\dm^\ite{t+1}$ using the parameter at iteration $t$ ($\mm = \mm^\ite{t},
\Cinv = ((\sigma^\ite{t})^2 \cov^\ite{t})^{-1}$), as follows:
\begin{align}
  \dm^\ite{t+1} = \frac{\norm{\mm - \tilde{\mm}}^2_\Cinv \cdot \hat{\sigma}^2}{n (n / 2 + \cact)}
  \exp \left( \frac{\min(0, \lambda_\mathrm{def} - \lambda)}{\lambda} \right)
  \enspace,
  \label{eq:dm}
\end{align}%
where $\cact = | \set{j \mid g_j(\tilde{\mm}) = 0, \ \forall j} |$ is the number of
active constraints at the repaired mean $\tilde{\mm}$. From the perspective of
\eq{eq:snf} and \eq{eq:approx_mdist}, we wish to maintain $\dm \approx 1$ if the
denominator is $n^2$.  However, in \eq{eq:dm}, we replace $n^2$ with $n (n/2 +
\cact)$, leading to a variation in the denominator in $[n^2/2, 3n^2/2]$. The
motivation is to incorporate the fact that the numerator is expected to be smaller if
$\cact$ is smaller, as the projection is performed only on this $\cact$-dimensional
subspace.  Moreover, we introduce the exponential term to prevent the search
distribution from being unnecessarily biased toward the boundary if the population
size $\lambda$ is larger than the default $\lambda_\mathrm{def}$. If $\lambda \in
\Omega(n)$, $\hat{\sigma}$ will be in $\mathcal{O}(n)$, while $\hat{\sigma} \in
\mathcal{O}(\muw)$ if $\lambda \ll n$.  Then, to maintain $\dm \approx 1$, $\mm$ needs
to be closer to $\tilde{\mm}$, as $\lambda$ is greater.  That is, a larger $\lambda$
results in the search distribution being more biased toward the feasible domain, and
less efficient search performance.  To mitigate this problem, we allow $\mm$ to be up
to $e$ times away from $\tilde{\mm}$ by introducing the exponential term when
$\lambda$ is greater than $\lambda_\mathrm{def}$.  We adapt $\alpha$ so that $\dm$
will remain approximately 1, as follows:

\paragraph{Ranking Coefficient Adaptation}

Initialize $\alpha^\ite{0} = 1$, $\dm^\ite{0} = 0$.  At iteration $t$,
$\alpha^\ite{t}$ is updated as
\begin{align} \alpha^\ite{t+1} = \alpha^\ite{t} \cdot \left( \frac{\sgn (\dm^\ite{t+1}
-1)}{n} \right) \label{eq:alpha}
\end{align}%
only if $\sgn(\dm^\ite{t+1} - 1) = \sgn(\dm^\ite{t+1} - \dm^\ite{t})$ or
$\dm^\ite{t+1} = 0$, the latter of which is necessary to decrease $\alpha$ when the
mean vector remains in the feasible domain; that is, $\dm^\ite{t} = \dm^\ite{t+1} =
0$.  Following the update, $\alpha^\ite{t+1}$ is clipped to $[1/\lambda, \lambda]$,
because $\alpha < 1/\lambda$ and $\alpha > \lambda$ result in $\Rg$ and $\Rf$ being
ignored, respectively \footnote{We observed in preliminary experiments that the best
  result was obtained by using \cref{eq:alpha}.}.

\section{Invariance Properties of ARCH}
\label{sec:invariance}

We prove that ARCH is invariant to the problem transformations described in
\Cref{sec:guideline}.  ARCH receives the candidate solutions
$(\xx_k)_{k=1}^{\lambda}$, and distribution parameters $\mm$ and $\Sigma = \sigma^2
\CC$, from the CMA-ES, and returns the total rankings
$\big(\RT(\xx_k)\big)_{k=1}^{\lambda}$ of the candidate solutions to the CMA-ES.
Meanwhile, it maintains the penalty coefficient $\alpha$ and retains $\dm$ for the
next update.  Therefore, the functionality of ARCH can be formulated as
\begin{equation}
  ((\RT(\xx_k))_{k=1}^\lambda, \alpha^{(t+1)}, \dm^{(t+1)})
  =  \textsc{ARCH}((\xx_k)_{k=1}^\lambda, \mm, \Cinv, \alpha^{(t)}, \dm^{(t)}, F) \enspace,
  \label{eq:arch}
\end{equation}
where $F : \xx \mapsto (f(\xx), g_1(\xx), \dots,g_m(\xx))$ is the vector-valued
function consisting of the objective and constraints.  In the following, we prove that
the outputs of \eqref{eq:arch} remain unchanged by the abovementioned transformations.

\subsection{ARCH on Inner Product Space}

We begin by defining ARCH on an inner product space $(V,\ \inner{\cdot}{\cdot})$.

A random vector $X_V$ on the inner product space $(V,\ \inner{\cdot}{\cdot})$ has a
normal distribution $\mathcal{N}(\mu_V, \Sigma_V)$ if $\inner{\bm{v}}{X}$ has the
normal distribution $\mathcal{N}(\inner{\bm{v}}{\mu_V}, \inner{\bm{v}}{\Sigma_V
(\bm{v}}))$ for any $\bm{v} \in V$, where $\mu_V \in V$ is the mean vector and
$\Sigma_V: V \to V$ is the covariance \citep{eaton2007chapter}. Let $F^V = (f^V,
g_1^V, \dots, g_m^V)$ be the objective and constraint functions defined on $V$ as
\eqref{eq:vectorspace}. ARCH on the inner product space receives these distribution
parameters $\mu_V$ and $\Sigma_V$, candidate solutions $\pp_1,\dots,\pp_\lambda$ drawn
from the normal distribution, and the objective and constraint functions $F^V$.

Let $\pp$ and $\qq$ be the vectors in $V$. The Mahalanobis distance between $\pp$ and
$\qq$ under the covariance $\Sigma_V$ is defined as
\begin{align*}
  \textstyle \norm{\pp - \qq}_{\Sigma_V^{-1}}^2 := \textstyle \inner{\pp - \qq}{\Sigma_V^{-1}(\pp - \qq)} \enspace.
\end{align*}%
Let $\mathcal{S}^V = \set{\pp \in V \mid g_j^V(\pp) \leq 0, \ \forall j
\in \llbracket 1, m \rrbracket}$ be the feasible domain.  Given a solution $\pp \in
V$, let $\J^V(\pp) = \set{j \in \llbracket 1, m \rrbracket \mid g_j^V(\pp) > 0}$ be
the set of indices of the unsatisfied constraints, and let $\mathcal{A}^V(\pp) =
\set{\qq \in V \mid g_j^V(\qq) = 0, \ \forall j \in \J^V(\pp)}$ be the intersection of
the violated constraint boundaries. The repair operation is defined as follows:
\begin{align}
  \mathtt{Repair}^{V}(\pp) = \argmin_{\qq \in V} \ \norm{\pp - \qq}^2_{\Sigma^{-1}_{V}} \quad
  \text{s.t.\ } \ \qq \in
  \begin{cases}
    \mathcal{A}^V(\pp) \cap \mathcal{S}^V
    & \text{if $\mathcal{A}^V(\pp) \cap \mathcal{S}^V \neq \emptyset$}
    \\
    \mathcal{S}^{V}
    & \text{otherwise}
    \enspace .
  \end{cases}
      \label{eq:repair:vector}
\end{align}%
Let $\dm$ and $\alpha$ be computed as in \eqref{eq:dm} and \eqref{eq:alpha}, where
\del{%
  $\norm{\mm - \mathtt{Repair}(\mm)}_{\Cinv}^2$, and $\cact$ is replaced with
  $\norm{\mu_V - \mathtt{Repair}^{V}(\mu_V)}_{\Sigma_V^{-1}}^2$ and $\cact = | \set{j
    \mid g_j^V(\mathtt{Repair}^{V}(\mu_V)) = 0, \ \forall j} |$}{}%
\new{
  $\mm$, $\tilde{\mm} = \mathtt{Repair}(\mm)$, $\Cinv$, and $\cact$ are
  replaced with $\mu_V$, $\tilde{\mu}_V = \mathtt{Repair}^V(\mu_V)$,
  $\Sigma_V^{-1}$, and $c_{\rm act}^V = | \set{j \mid g_j^V(\tilde{\mu}_V) =
  0, \ \forall j} |$, respectively.
}Analogously to
\cref{eq:rt,eq:rf,eq:rg}, we compute
\begin{align}
  \RT^V(\pp_k) = \Rf^V(\pp_k) + \alpha \Rg^V(\pp_k)
  \enspace,
\end{align}%
where
\begin{align}
  \Rf^V(\pp_k) &= \textstyle \sum^{\lambda}_{l=1}\1{f^V(\tilde{\pp}_l) < f^V(\tilde{\pp}_k)}
                 + \frac{1}{2} \sum^{\lambda}_{l=1} \1{f^V(\tilde{\pp}_l) = f^V(\tilde{\pp}_k)} \\
  \Rg^V(\pp_k) &= \textstyle \sum^{\lambda}_{l=1} \1{\gsig^V(\pp_l) < \gsig^{V}(\pp_k)}
                 + \frac{1}{2} \sum^{\lambda}_{l=1} \1{\gsig^V(\pp_l) = \gsig^V(\pp_k)}
                 \enspace,
\end{align}%
in which $\gsig^V(\pp) = \norm{\pp - \tilde{\pp}}^2_{\Sigma^{-1}_V}$ and $\tilde{\pp}
= \mathtt{Repair}^{V}(\pp)$.

The operation of ARCH on the inner product space $(V,\ \inner{\cdot}{\cdot})$ is then
expressed as
\begin{equation}
  ((\RT^V(\pp_k))_{k=1}^\lambda, \alpha^{(t+1)}, \dm^{(t+1)})
  =  \textsc{ARCH}^V((\pp_k)_{k=1}^\lambda, \mu_V, \Sigma_V^{-1}, \alpha^{(t)}, \dm^{(t)}, F^V) \enspace.
  \label{eq:arch:vector}
\end{equation}
Note that $\textsc{ARCH}^V$ is defined without relying on any coordinate system.  This
implies that $\textsc{ARCH}^V$ is invariant to any coordinate system transformation.

\subsection{Invariance to Affine Transformation of Search Space Coordinates}

Firstly, we demonstrate that ARCH is invariant to an arbitrary affine transformation
of the search space coordinate system. For this purpose, it is sufficient to prove
that the operation of ARCH \eqref{eq:arch} on any given coordinate system $\xx =
\psi(\pp)$ is equivalent to the operation of ARCH \eqref{eq:arch:vector} on the inner
product space $(V,\ \inner{\cdot}{\cdot})$.

Note that the normal distribution $\mathcal{N}(\mu_V, \Sigma_V)$ on $V$ corresponds to
the normal distribution $\mathcal{N}(\mu_\psi, \Sigma_\psi)$ on $\R^n$ in the
coordinate system $\psi: V \to \R^n$, where $\mu_\psi = \psi(\mu_V)$ and
$[\Sigma_{\psi}]_{i,j} =\inner{\bm{e}_i}{ \Sigma_V ( \bm{e}_j )}$.  The objective and
constraint function $F:\R^n \to \R^{m+1}$ corresponding to $F^V:V \to \R^{m+1}$ is
expressed as $F = F^{V}\circ \psi^{-1}$.

\begin{theorem}
  \label{theorem:affine}
  Let $\psi: V \to \R^n$ be an arbitrary coordinate system.
  Let $\mu_V \in V$ be an arbitrary vector and $\Sigma_V: V \to V$ be an arbitrary positive definite symmetric linear transformation.
  Let $\mu_\psi = \psi(\mu_V)$ and $[\Sigma_{\psi}]_{i,j} = \inner{\bm{e}_i}{
    \Sigma_V ( \bm{e}_j )}$ and $F = F^V \circ \psi^{-1}$, assume that the feasible domain $\mathcal{S}$ is convex.
  Then, for any $\pp_k \in V$ for $k = 1,\dots, \lambda$, $\alpha > 0$ and $\dm > 0$,
  \begin{equation*}
    \textsc{ARCH}((\psi(\pp_k))_{k=1}^\lambda, \mu_\psi, \Sigma_\psi^{-1}, \alpha, \dm, F)
    = \textsc{ARCH}^V((\pp_k)_{k=1}^\lambda, \mu_V, \Sigma_V^{-1}, \alpha, \dm,  F^V) \enspace.
  \end{equation*}
\end{theorem}

The proof of \Cref{theorem:affine} is provided in \Cref{proof:affine}.

\subsection{Invariance to Element-wise Increasing Transformation}

Next, we demonstrate that ARCH is invariant to an arbitrary element-wise increasing
transformation. That is, the ranking and internal parameter updates are not affected
by the transformation.
\begin{theorem}
  \label{theorem:increasing}
  Let $H = (h_0, h_{1}, \dots, h_m)$ be an arbitrary element-wise increasing
  transformation, assume that the feasible domain $\mathcal{S}$ is convex.
  For any $\xx_k \in \R^n$ for $k = 1,\dots, \lambda$, $\alpha > 0$, and $\dm > 0$,
  \begin{equation*}
    \textsc{ARCH}((\xx_k)_{k=1}^\lambda, \mm, \Cinv, \alpha, \dm, F)
    = \textsc{ARCH}((\xx_k)_{k=1}^\lambda, \mm, \Cinv, \alpha, \dm, H\circ F) \enspace.
  \end{equation*}
\end{theorem}

The proof of \Cref{theorem:increasing} is provided in \Cref{proof:increasing}.

\subsection{Invariance Properties of Entire Algorithm}

Finally, we discuss the invariance properties of the search algorithm including ARCH.

Suppose that the underlying unconstrained optimization algorithm is defined on an
inner product space, and the following steps are repeated:
\begin{enumerate}\setlength\itemsep{0em}
\item $\pp_1, \dots, \pp_\lambda = \textsc{sample}(\mu_V, \Sigma_V, \theta_V)$;
\item $\Rf(\pp_1), \dots, \Rf(\pp_\lambda) = \textsc{evaluate}((\pp_k)_{k=1}^{\lambda}, f^V)$; and
\item $\mu_V, \Sigma_V, \theta_V = \textsc{update}((\pp_k, \Rf(\pp_k))_{k=1}^{\lambda}, \mu_V, \Sigma_V, \theta_V)$,
\end{enumerate}
where $\theta_V$ contains all of the information used in the algorithm.  By nature, it
is invariant to any strictly increasing transformation of the objective function
$f^V$, as the outputs of $\textsc{evaluate}$ are the rankings of the objective
function values, and it is invariant to any strictly increasing transformation
$h$. Moreover, as it is defined independently of a coordinate system, its
implementation under a given coordinate system $\psi$ produces $\psi(\pp_1), \dots,
\psi(\pp_\lambda), \mu_{\psi}, \Sigma_{\psi}, \theta_\psi$ at any iteration $t$,
where $\pp_1,\dots,\pp_\lambda$ are the candidate solutions mentioned above, while
$\mu_\psi$, $\Sigma_\psi$, and $\theta_\psi$ are the expressions of $\mu_V$,
$\Sigma_V$, and $\theta_V$, respectively, in the coordinate system $\psi$.

When solving an explicitly constrained minimization problem $F^V$, we simply replace
the evaluation step as follows:
\begin{enumerate}\setlength\itemsep{0em}
\item $\pp_1, \dots, \pp_\lambda = \textsc{sample}(\mu_V, \Sigma_V, \theta_V)$;
\item $\RT(\pp_1), \dots, \RT(\pp_\lambda), \alpha, \dm = \textsc{ARCH}^V((\pp_k)_{k=1}^\lambda, \mu_V, \Sigma_V^{-1}, \alpha, \dm, F^V)$; and
\item $\mu_V, \Sigma_V, \theta_V = \textsc{update}((\pp_k, \RT(\pp_k))_{k=1}^{\lambda}, \mu_V, \Sigma_V, \theta_V)$.
\end{enumerate}
As all of the operations are defined independently from a coordinate system, the
entire algorithm is invariant to any affine coordinate system
transformation. Moreover, because $\text{ARCH}^V$ is invariant to any element-wise
increasing transformation $H$, and the underlying algorithm only relies on the
rankings of the candidate solutions, all of the operations are invariant to $H$.

ARCH can be combined with a search algorithm that is not generalized to an inner
product space.  For example, our baseline CMA-ES, which is described in
\Cref{subsec:cmaes}, does not generalize to an inner product space (as $\ps$ is
coordinate dependent), although we empirically observe quite uniform behaviors under
different coordinate systems, as we observe in \Cref{sec:exp1}. ARCH does not disturb
the uniform behavior, as ARCH itself is invariant to any affine coordinate
transformation.

\section{Experiments on Linearly Constrained Quadratic Problems}
\label{sec:exp1}

Our first numerical experiments are aimed at demonstrating the effects of the
invariance to affine coordinate transformations of ARCH. We develop a set of
linearly constrained test problems. We compare ARCH with other CHTs for the
CMA-ES, demonstrating the manner in which affine coordinate transformations affect
the performances of CHTs that are not invariant thereto, while ARCH performs
equally effectively under different transformations. Moreover, we compare ARCH
with CHTs that are specialized for box constraints. Note that box CHTs are
expected to be more efficient for box-constrained optimization problems than other
CHTs for general linear constraints. We observe that ARCH is competitive with and
sometimes outperforms these, even on box-constrained problems.

\subsection{Linearly Constrained Quadratic Minimization Problems}
\label{subsec:test-problem}

We consider a linearly constrained problem (P0), defined in the $n$-dimensional inner
product space $(V, \inner{\cdot}{\cdot})$ on the real field $\R$ as
\eqref{eq:vectorspace}, where the constraints are defined as $g_i(\bm{p}) =
\inner{-\vj[i]}{\bm{p}} - [\lb]_i$ and $g_{n+i}(\bm{p}) = \inner{\vj[i]}{\bm{p}} -
[\ub]_i$, where $\bm{v}_i \in V$ is the normal vector of the $i$th constraint and
these are orthogonal to one another; that is, $\inner{\vj[i]}{\vj} = \delta_{i, j}$,
and $[\lb]_i,\ [\ub]_i \in \R$ are the lower and upper bounds, respectively.

We consider the coordinate system \del{$\psi(\pp) = \sum_{i=1}^{n} [
\bm{x}]_i \bm{v}_i$
with $\{ \vj[i] \}$ as the basis vectors. Let $\xx \in \R^n$ be a coordinate
vector.}{}%
\new{%
  $\psi: V \to \R^n$ with $\{ \vj[i] \}$ as the basis vectors.
  That is, $\psi : \bm{p} =  \sum_{i=1}^{n} [
\bm{x}]_i \bm{v}_i \mapsto \bm{x}$. 
}%
On this coordinate system, (P0) can be expressed as the box-constrained
minimization problem (P1), as follows:
\begin{align}
  \argmin_{\xx \in \R^n}\ f_{(P1)}(\xx) = f(\textstyle{\sum_{i=1}^{n} [\bm{x}]_i \bm{v}_i})
  \quad
  \text{s.t.\ }\ \bm{g}_{(P1)}(\xx) = \A \bm{x} - \bm{b} \preceq \bm{0}
    \enspace,
   \label{eq:p1}
\end{align}%
where $\A = [-\I, \I]^\T$ and $\bb = [\lb^\T, \ub^\T]$, and $\bm{g}_{(P1)} =
(g_1,\dots,g_{m})$ is a vector form of $m = 2n$ constraint functions. We define an
initial mean vector and initial covariance matrix as $\mm^\ite{0}$ and $\cov^\ite{0}$,
respectively, on this coordinate system.

By taking another basis $\{ \bm{w}_i \}$, the general linearly constrained
optimization problem (P2) can be defined as
\begin{align}
  \label{eq:p2}
  \argmin_{\bm{y} \in \R^n}\ f_{(P2)}(\bm{y}) = f_{(P1)}(\bm{P}\bm{y})
  \quad
  \text{s.t.\ }\ \bm{g}_{(P2)}(\bm{y}) = \A \bm{P}\bm{y} - \bm{b} \preceq \bm{0}
    \enspace.
\end{align}
On this coordinate system, the coordinate vector $\yy \in \R^n$ is transformed as $\xx
= \bm{P} \yy$ using a basis-transformation matrix $\bm{P}$, which transforms $\{
\vj[i] \}$ into $\{ \bm{w}_i \}$. The initial mean vector and covariance matrix are
transformed as $\bm{P}^{-1} \mm^\ite{0}$ and $\bm{P}^{-1} \cov^\ite{0}
(\bm{P}^{-1})^\T$, respectively.

These optimization problems (P1) and (P2) are equivalent to (P0). However, (P1) has a
box constraint, while (P2) has a set of linear constraints. Algorithms that are
invariant to any affine transformation of the search space must perform equivalently
on these problems. We use these problems to assess the invariance properties.

\begin{figure}[t]
  \begin{subfigure}{0.333\hsize}
    \centering
    \includegraphics[width=0.65\hsize]{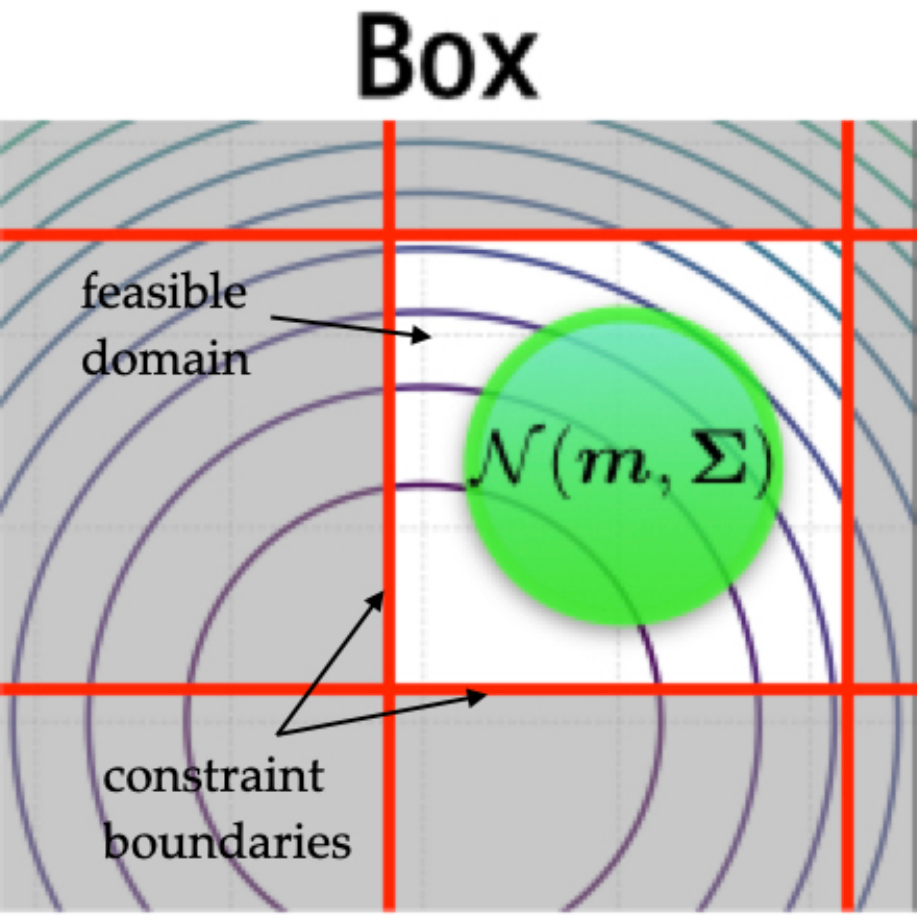}
  \end{subfigure}%
  \begin{subfigure}{0.333\hsize}
    \centering
    \includegraphics[width=0.65\hsize]{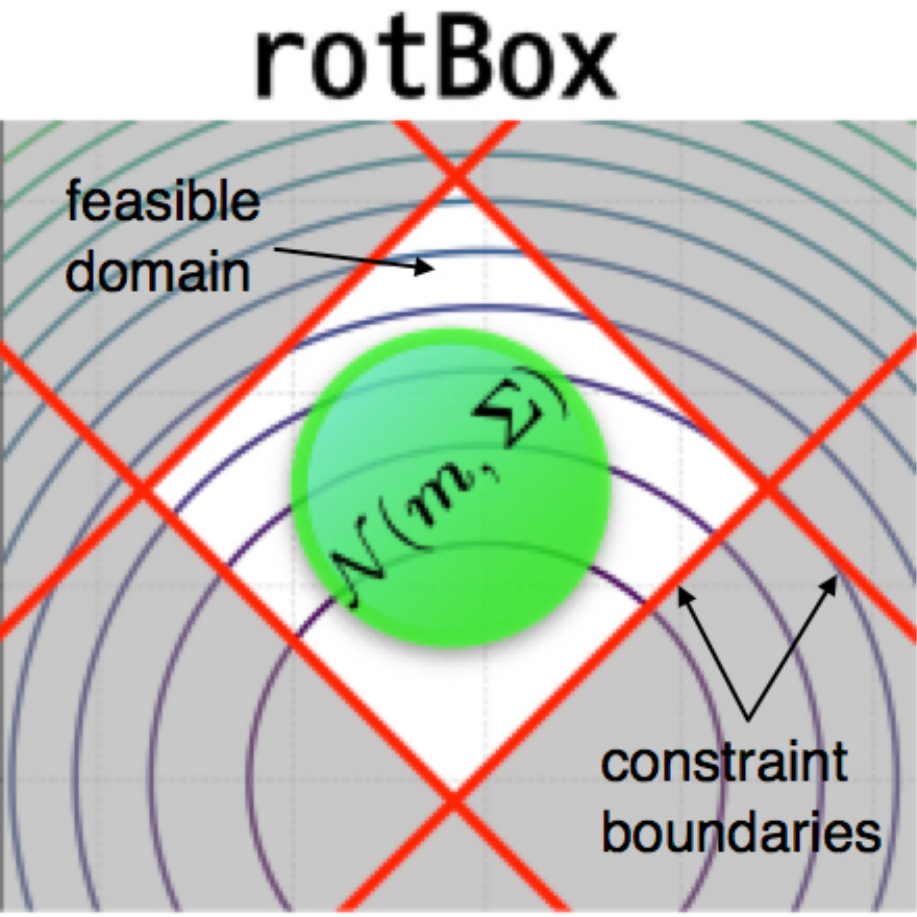}
  \end{subfigure}%
  \begin{subfigure}{0.333\hsize}
    \centering
    \includegraphics[width=0.65\hsize]{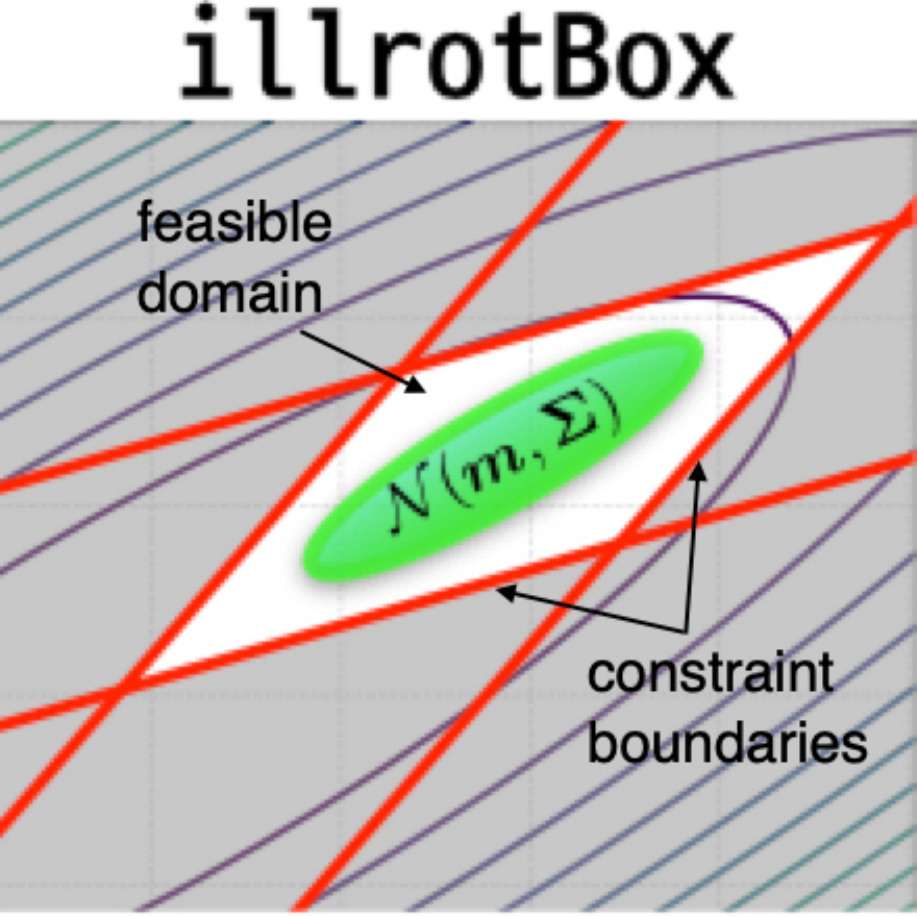}
  \end{subfigure}%
  \caption{Same problem with different coordinate systems.}
  \label{fig:constrained-problem}
\end{figure}

\subsection{Settings}
\label{subsec:settings}

For the test problem (P1) in \eqref{eq:p1}, we use three objective functions: sphere
($\fsph(\xx) = \sum^{n}_{i=1} [\xx]^2_i$), ellipsoid ($\fell(\xx) = \sum^{n}_{i=1}
10^{6 \frac{i-1}{n-1}} [\xx]^2_i$), and rotated ellipsoid ($\frotell(\xx) =
\fell(\bm{Q}_{\theta} \xx)$), where $\bm{Q}_{\theta} \in \R^{n \times n}$ is a block
diagonal matrix such that each block is a $2 \times 2$ orthogonal (that is, rotation)
matrix and all blocks share the same matrix. We let the counter-clockwise rotation
angle of each block be $\theta = \pi / 6$ in this experiment.

Problem (P1) is a box-constrained problem.  The lower and upper bounds are set as $\lb
= [-1, 1, \dots, -1, 1]^\T$ and $\ub = \lb + [5, \dots, 5]^\T$, respectively.  The
optimal solution, which is obtained by the Karush--Kuhn--Tucker (KKT) condition, is
located at $\xx^* = [0, 1, \dots, 0, 1]^\T$ for $\fsph$ and $\fell$, and $\xx^*
\approx [0.37, 1, \dots, 0.37, 1]^\T$ for $\frotell$.  That is, the even-numbered
coordinates of the optimum are on the boundary, while the others are not.  The number
of active constraints at the optimum is $n / 2$.

We use the following two matrices for the transformation matrix $\bm{P}$ in
\eqref{eq:p2}: $\Prot = \bm{Q}_{\theta'}$ and $\Pillrot = \bm{Q}_{\theta'}^\T \bm{D}
\bm{Q}_{\theta'}$ with $\theta' = \pi / 4$, where $\bm{Q}_{\theta'}$ is as defined
above, with $\theta \neq \theta'$, and $\bm{D} = \diag(1, 10, \dots, 1, 10)$ is a
diagonal matrix. The transformation matrix $\Prot$ only rotates the search space,
while $\Pillrot$ transforms the rectangular feasible domain into an $n$-parallelotope
shape.  Hereunder, problem (P1) is denoted by $\BOX$, the linearly constrained problem
(P2) using $\Prot$ is denoted by $\rotBox$, and (P2) using $\Pillrot$ is denoted by
$\illrotBox$. Their feasible domains and the levelsets of the objective function are
illustrated in \Cref{fig:constrained-problem}.

For the box-constrained optimization problem (P1), the initial mean vector is
$\mm^\ite{0} = \frac{\ub + \lb}{2} + \mathcal{U}(-1, 1)^n$, and the initial covariance
matrix is $\cov^\ite{0} = \I$. For the linearly constrained optimization problem (P2),
$\mm^\ite{0}$ and $\cov^\ite{0}$ are transformed in the manner described in
\Cref{subsec:test-problem}. The search space dimension is $n \in \{ 20, 50 \}$, the
initial step size is $\sigma^\ite{0} = \frac{1}{n} \sum^n_{i=1} \frac{[\ub]_i -
[\lb]_i}{4} = 1.25$, and the other parameters are set to their default values, as
defined in \Cref{tab:defaultparameter}.

\providecommand{\medfigsize}{0.33\hsize}
\providecommand{\submedfigsize}{1.\hsize}

\begin{figure}[t]
  \centering
  \begin{subfigure}{\medfigsize}
    \centering
    \includegraphics[width=\submedfigsize]{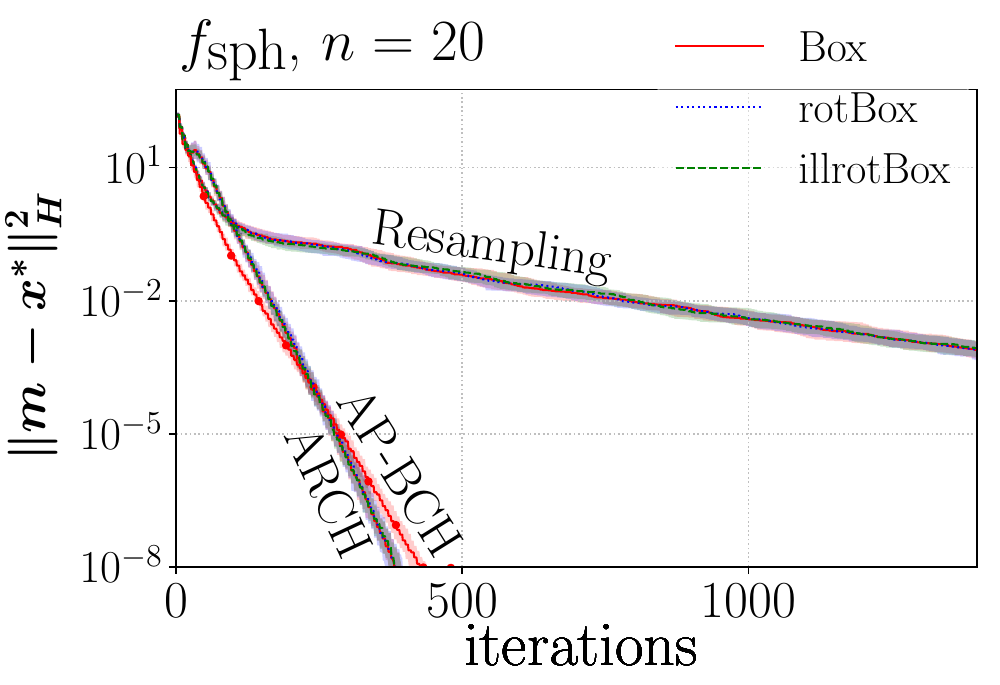}
  \end{subfigure}%
  \begin{subfigure}{\medfigsize}
    \centering
    \includegraphics[width=\submedfigsize]{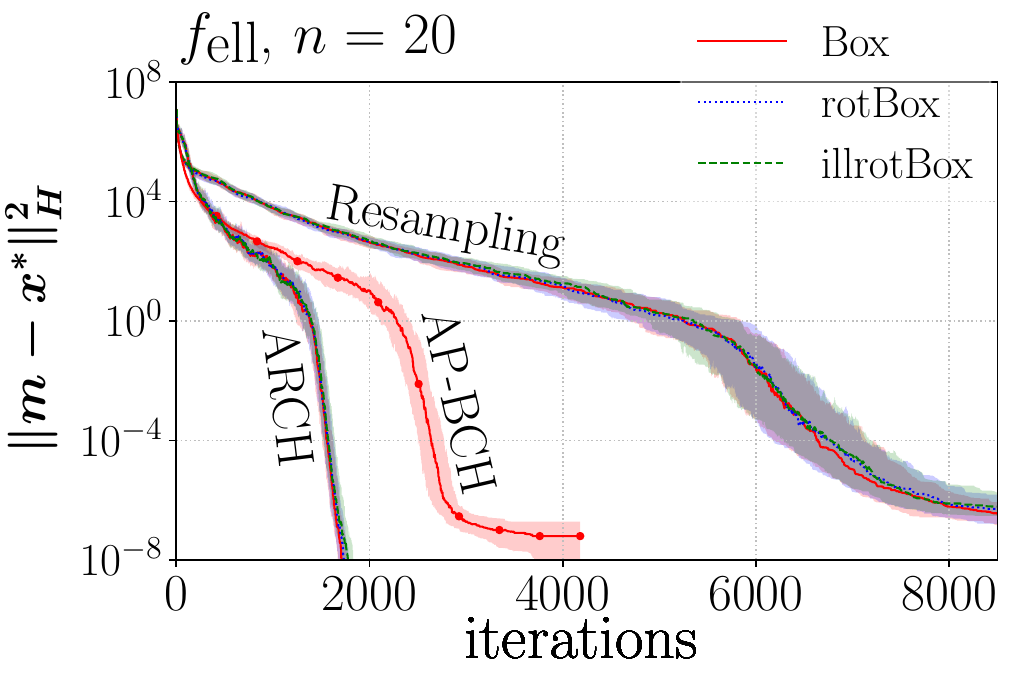}
  \end{subfigure}%
  \begin{subfigure}{\medfigsize}
    \centering
    \includegraphics[width=\submedfigsize]{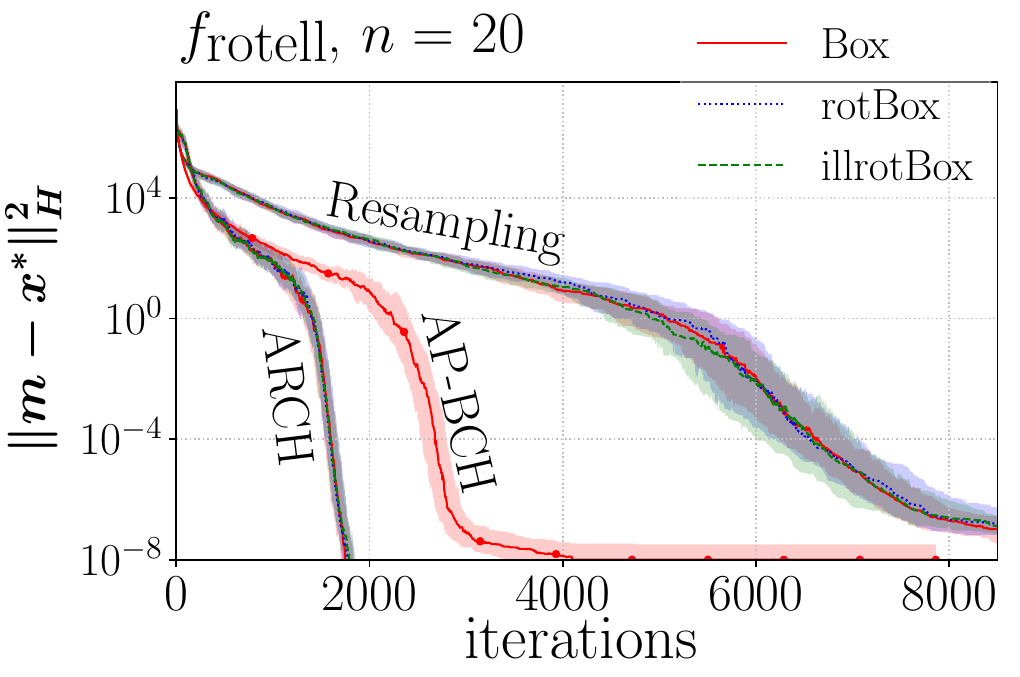}
  \end{subfigure}%
  \\
  \begin{subfigure}{\medfigsize}
    \centering
    \includegraphics[width=\submedfigsize]{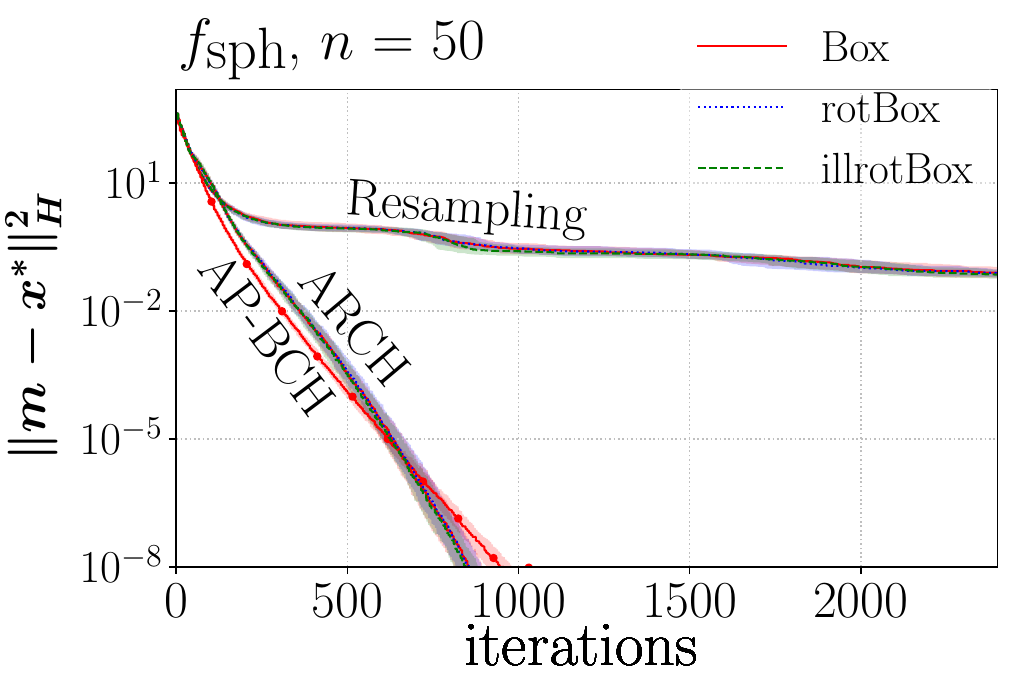}
  \end{subfigure}%
  \begin{subfigure}{\medfigsize}
    \centering
    \includegraphics[width=\submedfigsize]{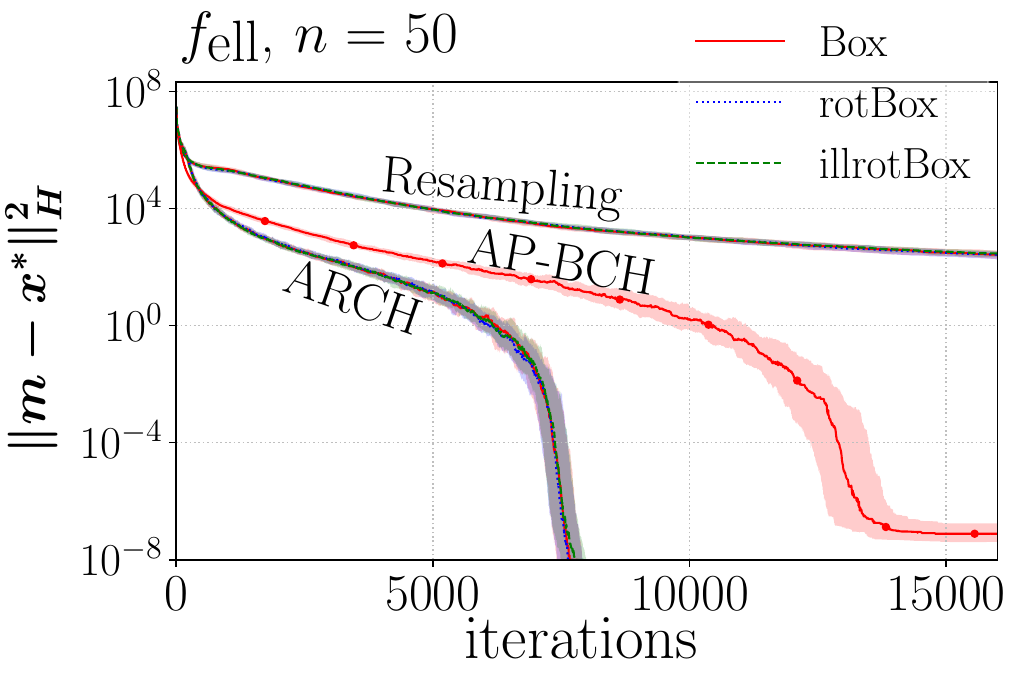}
  \end{subfigure}%
  \begin{subfigure}{\medfigsize}
    \centering
    \includegraphics[width=\submedfigsize]{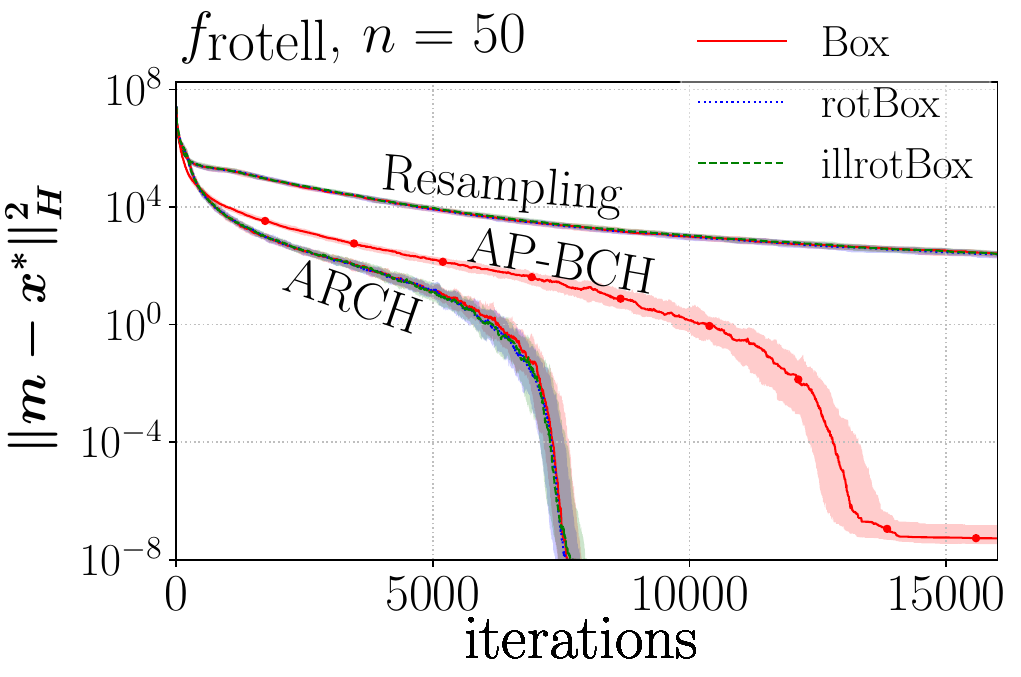}
  \end{subfigure}%
  \caption{Median (line) and 25\% to 75\%-ile range (band) over 100 trials.}
  \label{fig:medfig}
\end{figure}

\providecommand{\singlefigsize}{0.5\hsize}
\providecommand{\subsinglefigsize}{0.99\hsize}

\begin{figure}[t!]
  \centering
  \begin{subfigure}{\singlefigsize}%
    \centering%
    \includegraphics[width=\subsinglefigsize]{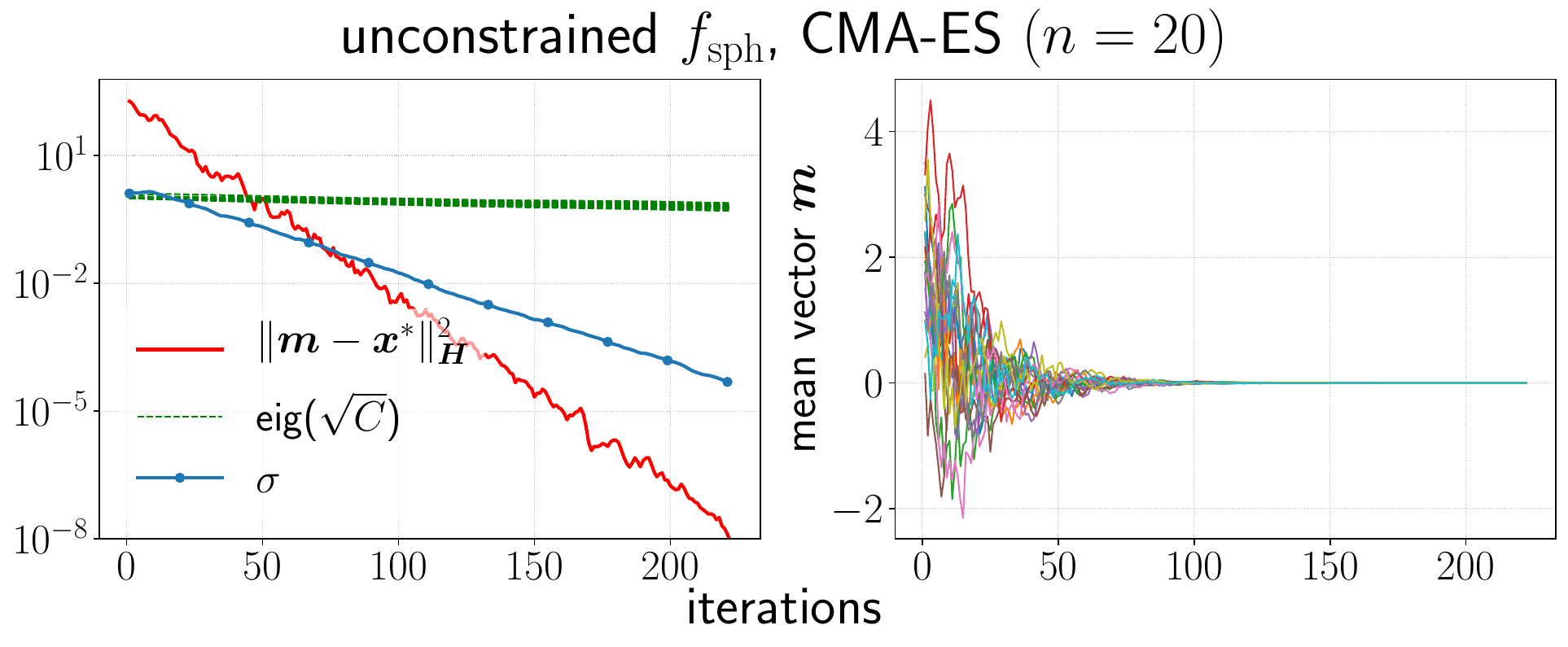}%
  \end{subfigure}%
  \begin{subfigure}{\singlefigsize}%
    \centering%
    \includegraphics[width=\subsinglefigsize]{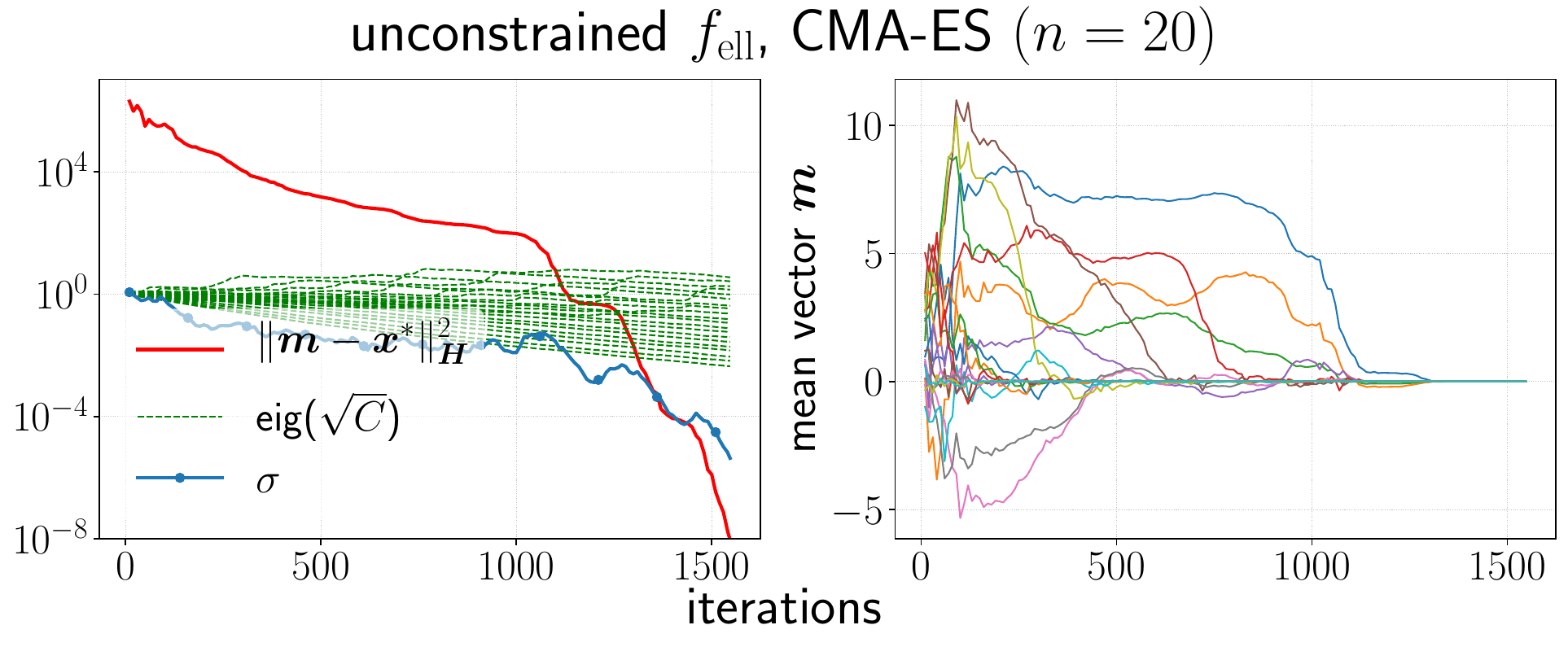}%
  \end{subfigure}%
  \\
  \begin{subfigure}{\singlefigsize}%
    \centering%
    \includegraphics[width=\subsinglefigsize]{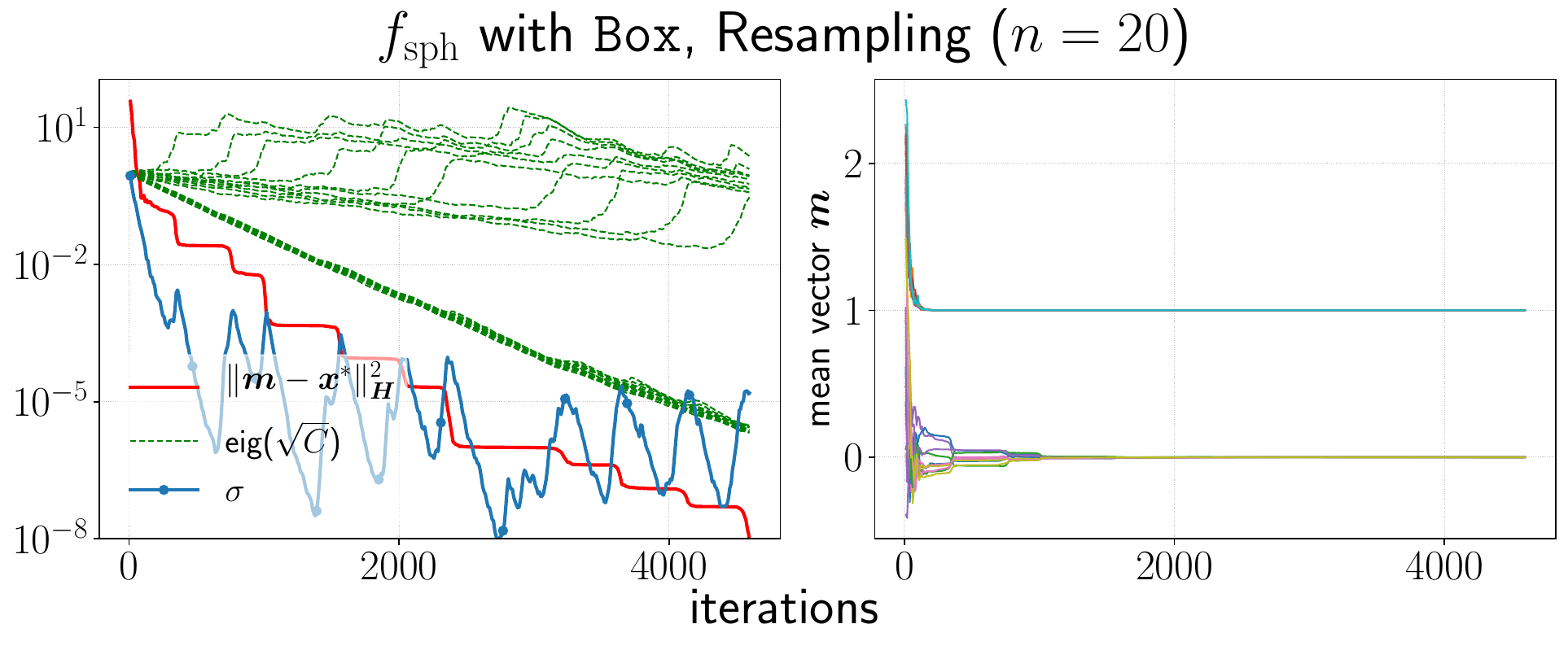}%
  \end{subfigure}%
  \begin{subfigure}{\singlefigsize}%
    \centering%
    \includegraphics[width=\subsinglefigsize]{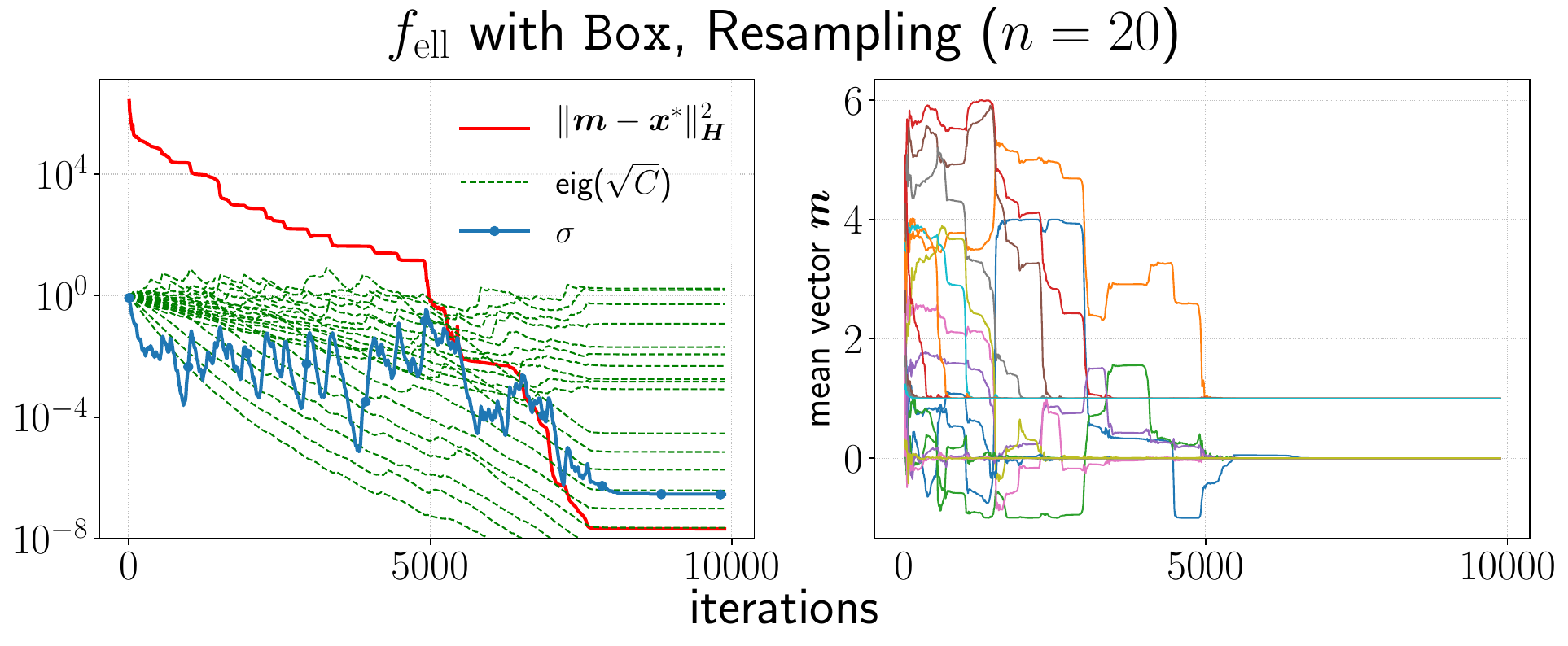}%
  \end{subfigure}%
  \\
  \begin{subfigure}{\singlefigsize}%
    \centering%
    \includegraphics[width=\subsinglefigsize]{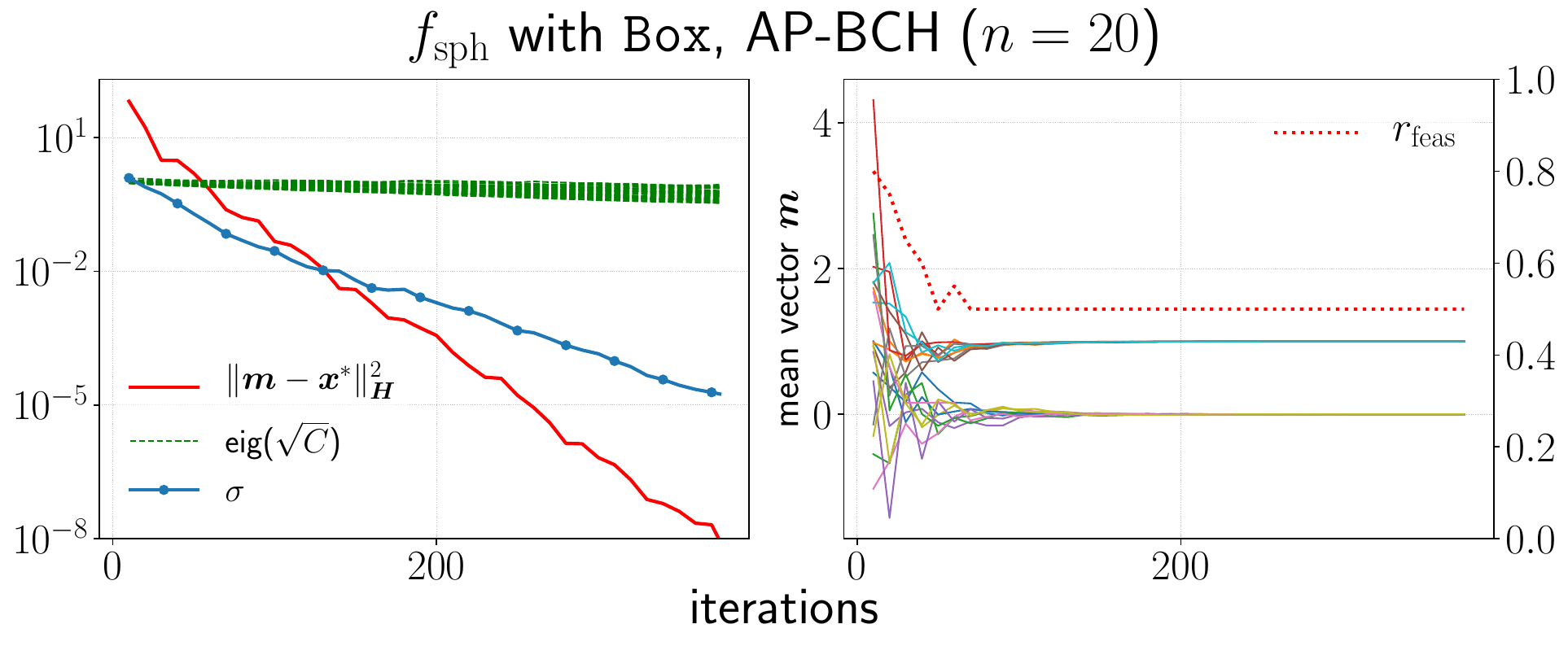}%
  \end{subfigure}%
  \begin{subfigure}{\singlefigsize}%
    \centering%
    \includegraphics[width=\subsinglefigsize]{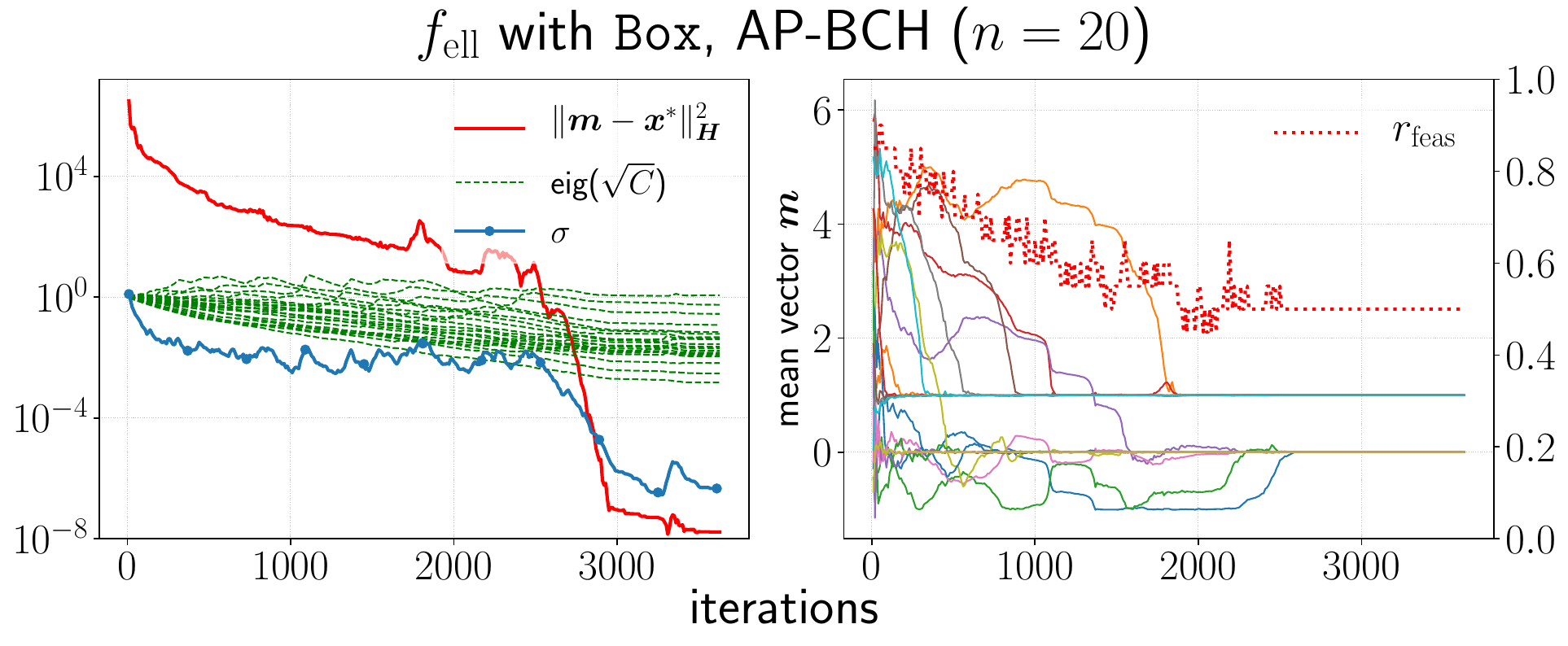}%
  \end{subfigure}%
  \\
  \begin{subfigure}{\singlefigsize}%
    \centering%
    \includegraphics[width=\subsinglefigsize]{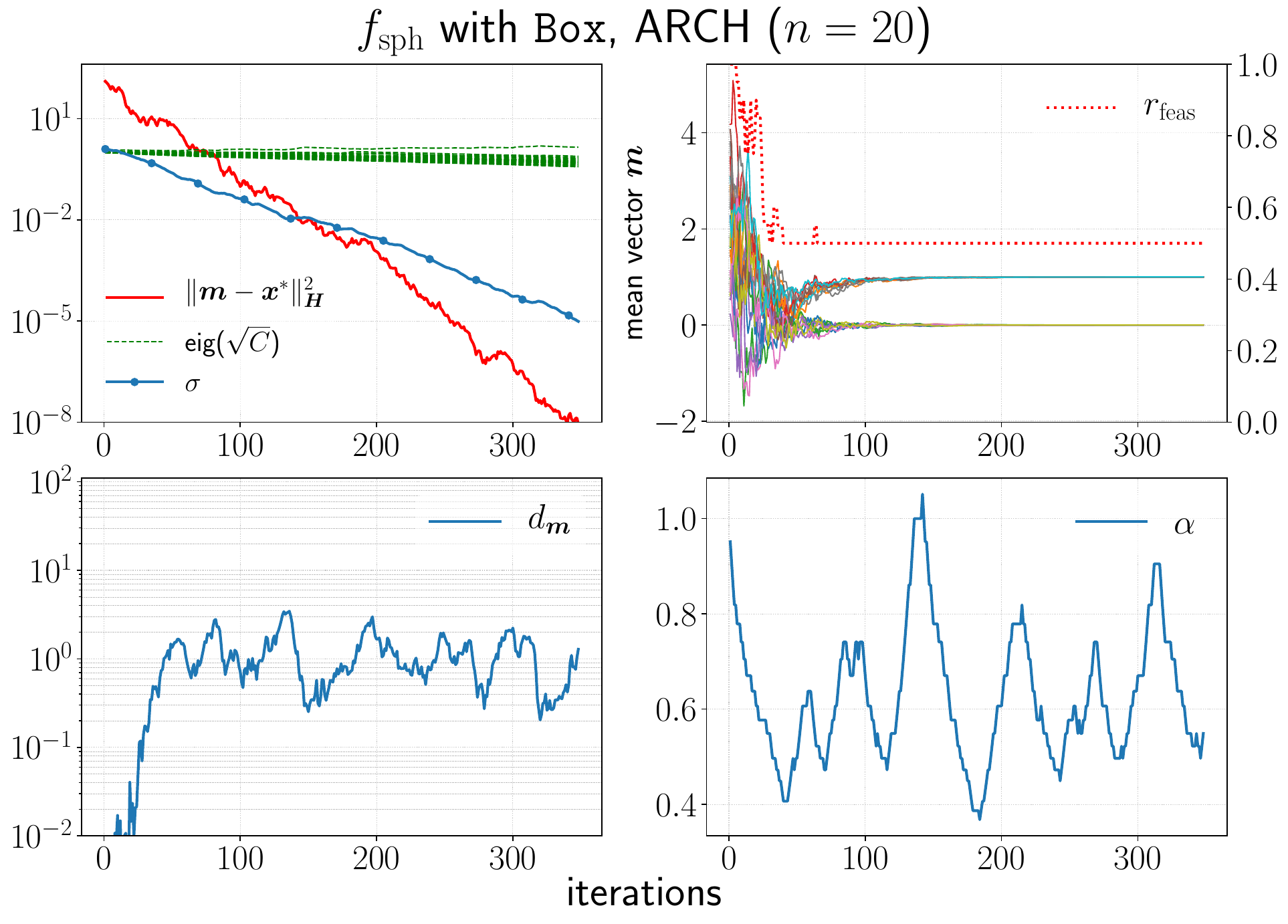}%
  \end{subfigure}%
  \begin{subfigure}{\singlefigsize}%
    \centering%
    \includegraphics[width=\subsinglefigsize]{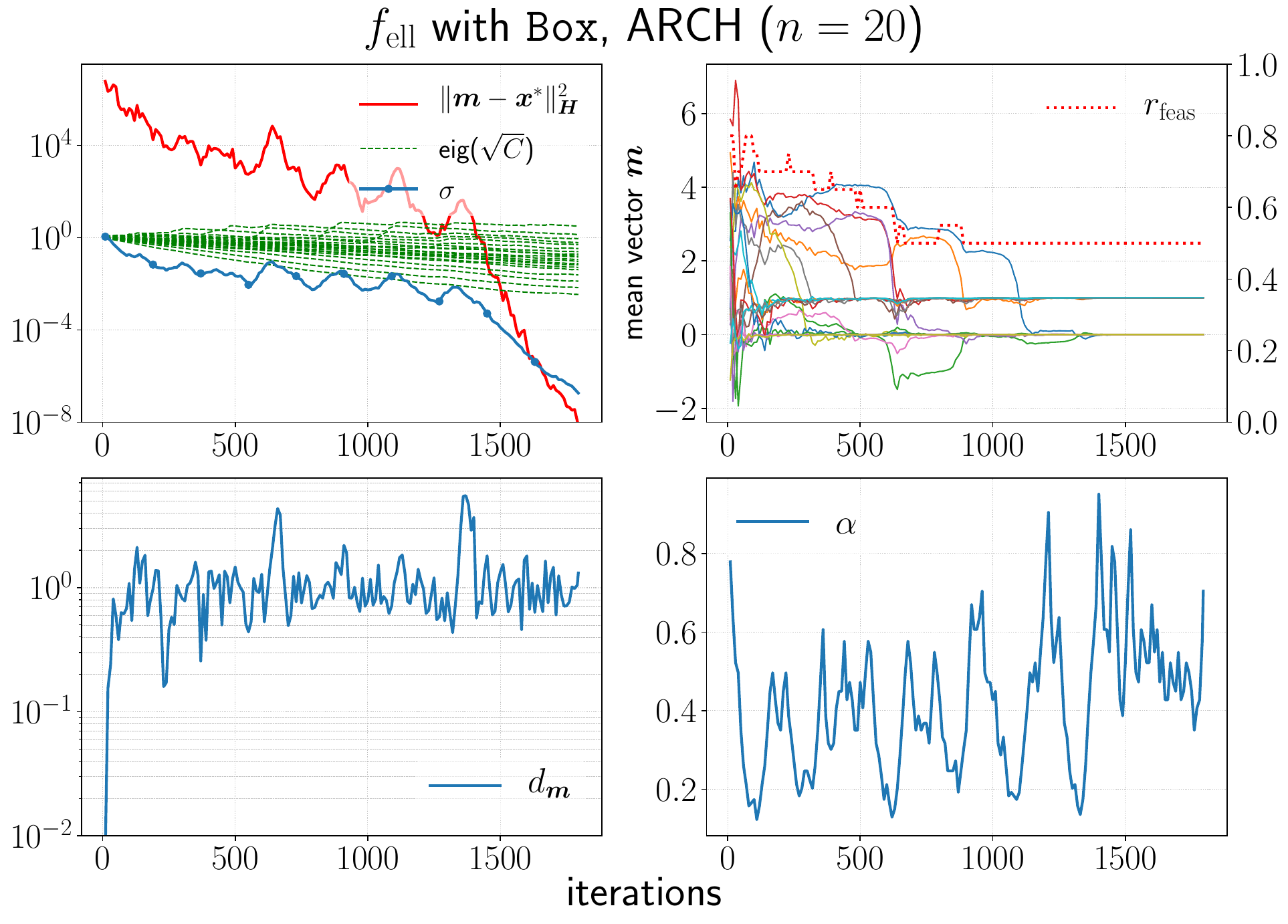}%
  \end{subfigure}%
  \caption{Typical single runs of CMA-ES (first row) on unconstrained $\fsph$ and
    $\fell$, resampling (second row), AP-BCH (third row), and ARCH (fourth row) on
    $\fsph$ and $\fell$ with $\BOX$. The figures indicate: the Mahalanobis distance
    $\criterion$ between the mean vector $\mm$ and optimal solution $\x^*$, given the
    Hessian matrix $\HH$ of the objective $f$; the step size $\sigma$; the eigenvalues
    $\text{eig}(\sqrt{\cov})$ of the square root of $\cov$; the coordinates of $\mm$;
    the ratio $\rfeas$ of the constraints satisfied by $\mm$; the coefficient $\alpha$
    and the parameter $\dm$ used for updating $\alpha$ versus the number of
    iterations.}
  \label{fig:single}
\end{figure}

\subsection{Results and Discussion}
\label{subsec:result}

\Cref{fig:medfig} presents the performance of ARCH, the resampling technique with a
maximum resampling number of 500, and AP-BCH~\citep{Hansen2009tec} on $\fsph$,
$\fell$, and $\frotell$ under three different coordinate systems ($\BOX$, $\rotBox$,
and $\illrotBox$) on $n = 20$ and $n = 50$ dimensions.  As AP-BCH is a box CHT, the
results are illustrated only for $\BOX$.  The optimization progress is measured by the
Mahalanobis distance between the mean vector and optimal solution $\criterion = (\mm -
\xx^*)^\T \HH (\mm - \xx^*)$ given the Hessian matrix $\HH \in \R^{n \times n}$ of the
objective function.  \Cref{fig:single} presents the results of typical runs of the
CMA-ES on unconstrained problems, resampling, AP-BCH, and ARCH on $\fsph$ and $\fell$
under the $\BOX$ constraint on $n = 20$ dimensions.  For AP-BCH and ARCH, the ratio of
the number of constraints satisfied by the mean vector $\mm$,
\begin{align} \rfeas = \frac{1}{n} \left| \Set{ i \in \llbracket 1, n \rrbracket |
[\lb]_i \leq [\mm]_i \leq [\ub]_i } \right| \enspace,
\end{align} is illustrated for discussion.

Firstly, we focus on the results of ARCH and the resampling technique in
\Cref{fig:medfig}.  Although the underlying CMA-ES is not mathematically proven to
be invariant to affine transformation of the search space, we observe in
\Cref{fig:medfig} that the lines of ARCH for $\BOX$, $\rotBox$, and $\illrotBox$
overlap one another, owing to the invariance of ARCH to the affine transformation
of the search space. \del{The resampling is}{}\new{For resampling as well, we observe} the same. This indicates
that the performance comparison can be conducted on the most convenient case,
namely the $\BOX$ constraint case. In the following, we focus on analyzing the
results on $\BOX$, which can be generalized to general cases.

Next, we investigate the behavior of ARCH. We observe in \Cref{fig:single} that
the parameter $\dm$ is maintained at approximately $1$. This is the desired
behavior, as we design the adaptation of $\alpha$ to maintain $\dm \approx 1$.
Moreover, we can observe from the results that the behaviors of the CMA-ES on the
unconstrained problem and the behaviors of the CMA-ES with ARCH and AP-BCH are
similar, in that the covariance matrix tends to be proportional to \nnew{the
  inverse of} the Hessian
matrix of the objective function. However, in \Cref{fig:medfig}, we observe the
difference between these two algorithms in terms of the adaptation speed of the
covariance matrix. On $\fell$ and $\frotell$, the CMA-ES with ARCH adapts the
covariance matrix significantly faster than that with AP-BCH. By adapting the
coefficient $\alpha$, ARCH appears to resemble the selection of candidate
solutions on an unconstrained problem better than AP-BCH.

The resampling technique exhibits different behavior. The eigenvalues of the
covariance matrix are divided into two, with the smaller values corresponding to
the axes where the constraints are active at the optimum, and the greater values
corresponding to the axes where the constraints are inactive at the optimum. If we
do not use any information from the infeasible domain and generate only feasible
candidate solutions, this is a reasonable approach to cause the distribution to be
narrow in the directions of the active constraints. This concept has indeed been
employed in \citep{Arnold:2012:gecco}. However, this strategy significantly
reduces the speed of the approach of the mean vector to the optimum on the
boundary.

We provide additional experimental results in \Cref{sec:lin-prob-general}.

\section{Experiments on CEC 2006 Constrained Optimization Testbed}
\label{sec:exp2}

The aim of our second numerical experiments is twofold.

Firstly, we demonstrate that ARCH can be applied to nonlinearly as well as linearly
constrained problems.  To evaluate the efficacy, we compare ARCH with the active-set
ES~\citep{Arnold2017gecco} that can deal with explicit (a priori) and nonlinear
constraints, because there is currently only one CHT designed for such a constraint.
This is the $(1+1)$-ES-based approach, and the covariance matrix adaptation is not
incorporated.

Secondly, we \ddel{prove}{}\nnew{show} that, even if
the infeasible solutions can be evaluated on the objective, ARCH
\nnew{, which only evaluates the solutions in the feasible domain,}
exhibits advantages in solving constrained optimization with explicit
(a priori) constraints. For this purpose, we compare ARCH with the
AL~\citep{atamna2016ppsn} and MCR~\citep{garcia2017mcr}, which are designed for
simulation-based constraints under the relaxed assumption that any infeasible
solutions can be evaluated on the objective.

\subsection{CEC 2006 testbed}

To compare the CHTs, we use the CEC 2006 testbed~\citep{Liang2006problem}, which
consists of 24 constrained problems including linear/nonlinear and equality/inequality
constraints, where the equality constraints $h_j(\xx) = 0$ are transformed into
inequalities, as described below \cref{eq:orig-prob}, with $\epsilon_\text{eq} =
10^{-4}$.  We reuse the CEC2006 testbed since this testbed is widely used in existing
works, which make it easy to compare algorithms.  All optimization variables are
bounded as $[\lb]_i \leq [\xx]_i \leq [\ub]_i$ ($i \in \set{1, \dots, n}$).  The bound
constraints are transformed into a set of linear inequality constraints, as in
\cref{eq:p1}.  Note that ARCH is independent of the manner in which they are
transformed, but certain existing CHTs, such as MCR, are dependent on this.
\Cref{tab:cec2006} summarizes the features of the constrained problems.  Our
implementation of the CEC2006 testbed is available in the repository
(https://github.com/naoking158/ARCH).

\begin{table}[t]
  \centering
  \caption{Details of 24 constrained problems of CEC 2006 testbed: $n$ is the search
space dimension; LI, NI, LE, and NE are the number of linear/nonlinear inequality
constraints and linear/nonlinear equality constraints, respectively; and
$m_\mathrm{act}$ is the number of active constraints at the optimum.}
  \label{tab:cec2006}
  \small
  \begin{tabular}{|c|c|c|c|c|c|c|c|}
    \bottomrule
    \textbf{Prob.} & $\bm{n}$ & \textbf{Type of} $\bm{f}$ & \textbf{LI} & \textbf{NI} & \textbf{LE} & \textbf{NE} & $\bm{m_\mathrm{act}}$ \\ \toprule\bottomrule
    g01   & 13  & quadratic   & 9  & 0  & 0  & 0  & 6    \\ \hline
    g02   & 20  & nonlinear   & 0  & 2  & 0  & 0  & 1    \\ \hline
    g03   & 10  & polynomial  & 0  & 0  & 0  & 1  & 1    \\ \hline
    g04   & 5   & quadratic   & 0  & 6  & 0  & 0  & 2    \\ \hline
    g05   & 4   & cubic       & 2  & 0  & 0  & 3  & 3    \\ \hline
    g06   & 2   & cubic       & 0  & 2  & 0  & 0  & 2    \\ \hline
    g07   & 10  & quadratic   & 3  & 5  & 0  & 0  & 6    \\ \hline
    g08   & 2   & nonlinear    & 0  & 2  & 0  & 0  & 0    \\ \hline
    g09   & 7   & polynomial  & 0  & 4  & 0  & 0  & 2    \\ \hline
    g10   & 8   & linear      & 3  & 3  & 0  & 0  & 6    \\ \hline
    g11   & 2   & quadratic   & 0  & 0  & 0  & 1  & 1    \\ \hline
    g12   & 3   & quadratic   & 0  & 1  & 0  & 0  & 0    \\ \hline
    g13   & 5   & nonlinear    & 0  & 0  & 0  & 3  & 3    \\ \hline
    g14   & 10  & nonlinear   & 0  & 0  & 3  & 0  & 3    \\ \hline
    g15   & 3   & quadratic   & 0  & 0  & 1  & 1  & 2    \\ \hline
    g16   & 5   & nonlinear   & 4  & 34 & 0  & 0  & 4    \\ \hline
    g17   & 6   & nonlinear   & 0  & 0  & 0  & 4  & 4    \\ \hline
    g18   & 9   & quadratic   & 0  & 13 & 0  & 0  & 6    \\ \hline
    g19   & 15  & nonlinear   & 0  & 5  & 0  & 0  & 0    \\ \hline
    g20   & 24  & linear      & 0  & 6  & 2  & 12 & 16   \\ \hline
    g21   & 7   & linear      & 0  & 1  & 0  & 5  & 6    \\ \hline
    g22   & 22  & linear      & 0  & 1  & 8  & 11 & 19   \\ \hline
    g23   & 9   & linear      & 0  & 2  & 3  & 1  & 6    \\ \hline
    g24   & 2   & linear      & 0  & 2  & 0  & 0  & 2    \\ \toprule
  \end{tabular}
\end{table}

\subsection{ARCH vs Active-Set ES}
\label{ssec:vs-active-set}

In this section, we compare ARCH with the active-set ES~\citep{Arnold2017gecco}.

\subsubsection{Settings}

We follow the experimental setup of \citet{Arnold2017gecco}.  Note that the test
problems used in \citet{Arnold2017gecco}, known as the Michalewicz/Schoenauer test
set~\citep{michalewicz1996evolutionary}, are equivalent to the first 11 constrained
problems in the CEC 2006 testbed.  The initial mean vector $\mm^\ite{0}$ is sampled
from the uniform distribution $\mathcal{U}[\lb, \ub]$, and then projected onto the
boundary of the feasible domain by the repair operator defined by each CHT if the
sampled point is infeasible.  The initial step size and covariance matrix are set to
$\sigma^\ite{0} = 0.2 \min \set{\ub - \lb}$ and $\cov^\ite{0} = \I$, respectively.  A
run is terminated if the algorithm reaches 1200 iterations or locates a feasible
candidate solution $\xx$ with the objective value $f(\xx) < f^* + \eps \abs{f^*}$,
where $f^*$ is the optimal objective function value, as reported in
\citet{Liang2006problem}, and $\eps \in \set{10^{-4}, 10^{-8}}$ is referred to as the
target accuracy.  The other parameters of the CMA-ES are set to the default values
listed in \Cref{tab:defaultparameter}.

\subsubsection{Results and Discussion}

\begin{table}[]
  \centering
  \caption{
    Median number of function evaluations (NFES), iterations (\# Iterations) and
    success rate (SR) among 100 independent runs for each problem and each
    algorithm. The results of the active-set ES were obtained from
    \citep{Arnold2017gecco}.}
  \label{tab:vs-active}
  \footnotesize
  \begin{tabular}{|c|c||c|c||c|c|c|}
    \bottomrule
    \multirow{2}{*}{Prob.} & Target Accuracy & \multicolumn{2}{c||}{\textbf{Active-set ES}} & \multicolumn{3}{c|}{\textbf{ARCH}}  \\ \cline{3-7}
                           & $\eps$ & \multicolumn{1}{c|}{NFES (\# Iterations)} & \multicolumn{1}{c||}{SR} & \multicolumn{1}{c|}{NFES} & \multicolumn{1}{c|}{\# Iterations} & \multicolumn{1}{c|}{SR} \\
    \toprule
    \bottomrule
    \multirow{2}{*}{$\texttt{g01}$}  & $10^{-4}$   & 30              & 61\%    & 154   & $\mathbf{14}$     & 96\%   \\ \cline{2-7}
                                     & $10^{-8}$   & 30              & 62\%    & 154   & $\mathbf{14}$     & 96\%   \\ \hline
    \multirow{2}{*}{$\texttt{g02}$}  & $10^{-4}$   & --              & 0\%     & --    & --                & 0\%    \\ \cline{2-7}
                                     & $10^{-8}$   & --              & 0\%     & --    & --                & 0\%    \\ \hline
    \multirow{2}{*}{$\texttt{g03}$}  & $10^{-4}$   & 463             & 100\%   & 900   & $\mathbf{90}$     & 100\%  \\ \cline{2-7}
                                     & $10^{-8}$   & 863             & 100\%   & 1720  & $\mathbf{172}$    & 100\%  \\ \hline
    \multirow{2}{*}{$\texttt{g04}$}  & $10^{-4}$   & 22              & 100\%   & 176   & 22                & 100\%  \\ \cline{2-7}
                                     & $10^{-8}$   & 24              & 100\%   & 176   & $\mathbf{22}$     & 100\%  \\ \hline
    \multirow{2}{*}{$\texttt{g05}$}  & $10^{-4}$   & $\mathbf{36}$   & 100\%   & 624   & 78                & 97\%   \\ \cline{2-7}
                                     & $10^{-8}$   & $\mathbf{82}$   & 100\%   & 1160  & 145               & 41\%   \\ \hline
    \multirow{2}{*}{$\texttt{g06}$}  & $10^{-4}$   & 5               & 100\%   & 6     & $\mathbf{1}$      & 100\%  \\ \cline{2-7}
                                     & $10^{-8}$   & 5               & 100\%   & 6     & $\mathbf{1}$      & 100\%  \\ \hline
    \multirow{2}{*}{$\texttt{g07}$}  & $10^{-4}$   & 325             & 100\%   & 1635  & $\mathbf{163.5}$  & 100\%  \\ \cline{2-7}
                                     & $10^{-8}$   & 557             & 100\%   & 2705  & $\mathbf{270.5}$  & 100\%  \\ \hline
    \multirow{2}{*}{$\texttt{g08}$}  & $10^{-4}$   & 107             & 38\%    & 510   & $\mathbf{85}$     & 57\%   \\ \cline{2-7}
                                     & $10^{-8}$   & 210             & 44\%    & 636   & $\mathbf{106}$    & 57\%   \\ \hline
    \multirow{2}{*}{$\texttt{g09}$}  & $10^{-4}$   & 307             & 100\%   & 846   & $\mathbf{94}$     & 100\%  \\ \cline{2-7}
                                     & $10^{-8}$   & 582             & 100\%   & 1620  & $\mathbf{180}$    & 100\%  \\ \hline
    \multirow{2}{*}{$\texttt{g10}$}  & $10^{-4}$   & 117             & 100\%   & 580   & $\mathbf{58}$     & 100\%  \\ \cline{2-7}
                                     & $10^{-8}$   & $\mathbf{236}$  & 100\%   & 2985  & 298.5             & 100\%  \\ \hline
    \multirow{2}{*}{$\texttt{g11}$}  & $10^{-4}$   & 25              & 100\%   & 60    & $\mathbf{10}$     & 100\%  \\ \cline{2-7}
                                    & $10^{-8}$    & 73              & 100\%   & 258   & $\mathbf{43}$     & 100\%  \\
    \toprule
  \end{tabular}
\end{table}

We conduct 100 runs of each algorithm for each problem.  The results are summarized in
\Cref{tab:vs-active}.  The success rate is defined as the number of runs in which the
algorithm can reach the target within 1200 iterations, divided by the total number of
runs (100).

Comparing the median number of $f$-calls for reaching the same accuracy, $\epsilon$,
in \Cref{tab:vs-active}, we can observe that the active-set ES achieves the target
accuracy with a lower number of $f$-calls.  However, the median number of iterations
is lower for ARCH, as it is the population-based approach.  As the number of candidate
solutions per iteration is increased, we expect the number of iterations to decrease.
This is an advantage of ARCH when the candidate solutions can be evaluated on $f$ in
parallel.

Focusing on problems g01 and g08, ARCH exhibits higher success rates than the
active-set ES.  This is possibly because ARCH is combined with the $(\mu, \lambda)$
type ES, rather than the (1+1)-ES.  As these problems have local minima, ARCH can
reach the target at a higher rate than the active-set ES, which is (1+1)-ES based.
None of the algorithms can solve problem g02, because this problem has a strongly
multimodal landscape.

Another advantage that does not appear in this experiment is that we incorporate the
covariance matrix adaptation, which is desirable for ill-conditioned and non-separable
problems.

\subsection{ARCH vs AL \& MCR}
\label{ssec:vs-aal-mcr}

In this section, we compare ARCH with the AL~\citep{atamna2016ppsn} and
MCR~\citep{garcia2017mcr} techniques.

\subsubsection{Settings}

In this experiment, we use all 24 problems in the CEC 2006 testbed. Following the
default setup in \citet{Liang2006problem}, we regard each run as successful if a
feasible candidate solution $\xx$ with the objective function value $f(\xx) - f^* \leq
10^{-4}$ is located within $5 \times 10^5$ $f$-calls, where $f^*$ is the optimal
objective value.

Because we consider the constraints to be explicit, it is a natural approach to
prepare a set of feasible solutions and begin the optimization process with one
feasible solution as the initial search point. The set of feasible solutions is
generated as follows. We run the CMA-ES with the following loss function $L: \R^n \to
\R$,
\begin{align}
 L(\xx) = \sum^m_{j = 1} R_{g_j}(\xx) \enspace,
  \quad
  R_{g_j}(\xx) =
  \sum^\lambda_{k=1} \1{g_j^+(\xx_k) < g_j^+(\xx)}
  + \frac{1}{2} \sum^\lambda_{k=1} \1{g_j^+(\xx_k) = g_j^+(\xx)}
  \enspace,
\end{align}%
where $g_j^+(\xx) = \max(0, g_j(\xx))$.  If the candidate solutions are all feasible,
they receive the same ranking and the distribution parameter update becomes an
unbiased random walk.  Then, the CMA-ES tends to produce feasible solutions
extensively in a connected feasible subset.  To obtain diverse feasible solutions, we
run the CMA-ES until $10n$ feasible solutions are generated, and repeat it 50 times
with the following initial parameters: $\sigma^\ite{0} = \exp \big( \frac{1}{n}
\sum^n_{i=1} \ln \big( \frac{[\ub]_i - [\lb]_i}{5} \big) \big)$, $\cov^\ite{0} =
\diag(\frac{\ub - \lb}{5 \sigma^\ite{0}})^2$, and $\mm^\ite{0} \sim \mathcal{U}[\lb,
\ub]$.

Because multimodal functions exist in the CEC 2006 testbed, we use the BIPOP restart
strategy~\citep{Hansen2009geccobbobbi}, which updates the population size and initial
step size for every restart. Instead of randomly sampling an initial mean vector
$\mm^\ite{0}$ from the entire search space, we randomly sample one feasible solution
from the above prepared feasible set in each restart. The initial step size and
covariance matrix are set to $\sigma^\ite{0} = \exp \left( \frac{1}{n} \sum^n_{i=1}
\ln \left( \frac{[\ub]_i - [\lb]_i}{5} \right) \right)$ and $\cov^\ite{0} =
\diag(\frac{\ub - \lb}{5 \sigma^\ite{0}})^2$, respectively, with the step size scaled
by the BIPOP strategy in each restart. The other parameters of the CMA-ES are set to
the default values listed in \Cref{tab:defaultparameter}, and we follow the restart
condition described in \citet{Hansen2009geccobbobbi}.

\subsubsection{Results and Discussion}

All results are summarized in \Cref{tab:vs-qrsk}. The statistical significance is
tested using the two-sided Mann--Whitney rank test, with a significance level of $5\%
/ m$, where $m = 42$ is the number of tests (Bonferroni correction).
\footnote{
  \new{Since the tests have been performed excluding the results where no runs reached the target,
    the number of tests was 42, not $24 \times 3 = 72$.}
}

\begin{table}[t!]
  \centering
  \caption{Median number of function evaluations (NFES) and success rate (SR) among 25
    independent runs for each problem and each algorithm.  The subscript of the NFES
    indicates the average number of restarts for the successful runs.  The markers $*$ (AL
    vs ARCH), $+$ (AL vs MCR), and $\circ$ (MCR vs ARCH) indicate the statistical
    significance according to the two-sided Mann--Whitney rank test with a significance
    level of $5 / m \%$, where $m = 42$ is the number of tests (Bonferroni correction).
  }
  \label{tab:vs-qrsk}
  \begin{tabular}{|c||c|c||c|c||c|c|}
    \bottomrule
    \multirow{2}{*}{\textbf{Prob.}} & \multicolumn{2}{c||}{\textbf{AL}} & \multicolumn{2}{c||}{\textbf{MCR}} &  \multicolumn{2}{c|}{\textbf{ARCH}} \\ \cline{2-7}
                                    & \textbf{NFES} & \textbf{SR} & \textbf{NFES} & \textbf{SR} & \textbf{NFES} & \textbf{SR}  \\ \toprule\bottomrule
    g01 & -                         & 0\%   & $320245_{(4.92)}$   & 100\% & $\mathbf{737}_{(0.38)}^{\circ}$  & 100\%  \\ \hline
    g02 & -                         & 0\%   & -                     & 0\%   & -                       & 0\%    \\ \hline
    g03 & -                         & 0\%   & $10670_{(0.33)}$    & 100\% & $\mathbf{1030}_{(0.75)}^{\circ}$  & 100\%  \\ \hline
    g04 & -                         & 0\%   & $7728_{(0.25)}$     & 100\% & $\mathbf{120}_{(0.21)}^{\circ}$   & 100\%  \\ \hline
    g05 & $22056_{(10.00)}^{+}$ & 100\% & $269447_{(47.96)}$  & 8\%   & $\mathbf{737}_{(1.46)}^{\circ *}$  & 100\%  \\ \hline
    g06 & $8981_{(43.50)}$      & 4\%   & $1332_{(0.00)}^{+}$ & 100\% & $\mathbf{90}_{(3.62)}^{\circ *}$   & 100\%  \\ \hline
    g07 & $14415_{(2.42)}$       & 100\% & $19860_{(0.17)}$     & 100\% & $\mathbf{2782}_{(3.88)}^{\circ *}$ & 100\%  \\ \hline
    g08 & $468_{(0.46)}$         & 100\% & $162_{(0.17)}$  & 100\% & $390_{(3.08)}$       & 100\%  \\ \hline
    g09 & $2079_{(0.00)}^{+}$    & 100\% & $5328_{(0.00)}$     & 100\% & $1791_{(2.42)}^{\circ}$  & 100\%  \\ \hline
    g10 & -                         & 0\%   & $41780_{(1.12)}$     & 100\% & $\mathbf{2990}_{(0.33)}^{\circ}$  & 100\%  \\ \hline
    g11 & $294_{(0.00)}^{+}$    & 100\% & $3204_{(0.88)}$     & 100\% & $\mathbf{54}_{(0.17)}^{\circ *}$   & 100\%  \\ \hline
    g12 & $15601_{(6.08)}$      & 100\% & $2870_{(1.38)}^{+}$ & 100\% & $\mathbf{902}_{(6.21)}^{\circ *}$  & 100\%  \\ \hline
    g13 & $14208_{(2.12)}$       & 100\% & -                       & 0\%   & $\mathbf{2686}_{(4.96)}^{*}$  & 100\%  \\ \hline
    g14 & $418341_{(16.17)}$    & 12\%  & $16830_{(0.46)}^{+}$ & 100\% & $\mathbf{2172}_{(2.50)}^{\circ *}$ & 100\%  \\ \hline
    g15 & $2317_{(0.42)}^{+}$  & 100\% & $72239_{(11.17)}$     & 100\% & $\mathbf{175}_{(0.92)}^{\circ *}$  & 100\%  \\ \hline
    g16 & $5816_{(0.54)}$   & 100\% & $8664_{(0.46)}$       & 100\% & -                       & 0\%    \\ \hline
    g17 & $29400_{(5.92)}$      & 100\% & -                        & 0\%   & $\mathbf{1814}_{(4.17)}^{*}$  & 100\%  \\ \hline
    g18 & $11490_{(2.75)}$      & 100\% & $8810_{(1.58)}$       & 100\% & $\mathbf{3939}_{(6.83)}^{\circ *}$  & 100\%  \\ \hline
    g19 & -                        & 0\%   & $127308_{(1.38)}$     & 100\% & $\mathbf{5772}_{(0.08)}^{\circ}$  & 100\%  \\ \hline
    g20 & -                        & 0\%   & -                        & 0\%   & -                       & 0\%    \\ \hline
    g21 & -                        & 0\%   & -                        & 0\%   & -                       & 0\%    \\ \hline
    g22 & -                        & 0\%   & -                        & 0\%   & -                       & 0\%    \\ \hline
    g23 & -                        & 0\%   & $96078_{(34.38)}$     & 4\%   & $\mathbf{9386}_{(4.62)}^{\circ}$  & 100\%  \\ \hline
    g24 & $7422_{(3.33)}$      & 100\% & $600_{(0.21)}^{\bm{+}}$   & 100\% & $\mathbf{54}_{(0.38)}^{\circ *}$   & 100\% \\ \toprule
  \end{tabular}
\end{table}

The MCR and AL are CHTs that are often used for simulation-based constraints. They do
not exploit the fact that the constraints are explicit, but assume that $f$ is defined
on the infeasible domain.  With the exception of problems g08 and g16, ARCH
overwhelmingly outperforms the MCR and AL, which do not explicitly utilize the fact
that $g$-calls are computationally cheaper than $f$-calls, and perform as many
$f$-calls as $g$-calls. This indicates that ARCH can efficiently exploit the fact that
the constraints are explicit, even if the objective function is defined on the
infeasible domain.

ARCH does not reach the target for problems g02, g16, and g20 to g22.  For g02, g20,
g21, and g22, no algorithms can reach the target value.  Problem g02 has a strongly
multimodal landscape, and it is difficult to locate the global optimum among many
local optima.  Problems g20 to g22 have many nonlinear equality constraints, and it is
difficult to grasp the global landscape without using the function values outside of
the feasible domain. In particular, a feasible solution has not yet been determined
for problem g20 \citep{Liang2006problem}.  Problem g16 is a unique problem that can be
solved by the AL and MCR, but not ARCH.  In ARCH, the internal optimizer (SLSQP) used
in the repair operation fails to locate the solution.  This may occur if there are too
many complex constraints.

\section{Conclusions}\label{sec:conclusion}

A constrained continuous optimization problem has been addressed in this study, in
which the constraints are assumed to be explicitly written as mathematical expressions
and their evaluation time is negligible compared to that of the objective function. We
do not assume that the objective function values are defined outside of the feasible
domain. Our proposed CHT for the CMA-ES, known as ARCH, can handle explicit and
nonlinear constraints. ARCH is designed to be invariant to any element-wise
transformation of the objective and constraint functions, as well as to any affine
transformation of the search space coordinate, so as to preserve the invariance
properties of the underlying CMA-ES. The invariance properties are mathematically
proven and empirically validated. To the best of the authors' knowledge, this is the
first approach for explicit constraints that is invariant to those transformations.

Two sets of experiments revealed the effectiveness of ARCH. The first demonstrated
that \new{the convergence speed of} ARCH
\del{%
  is}{}%
\new{%
  improved compared with BCH%
}, in the especially ill-conditioned and non-separable objective function while it
is almost the same as in the well-conditioned problem.
\del{
  , more efficient than existing CHTs specialized for box constraints, even
  on box-constrained problems
}{}%
. These results could be attributed to the fact that
adapting the coefficient $\alpha$ allows for faster adaptation of the
covariance matrix $\cov$, and the affine invariance allows for a search behavior
that resembles one on the well-conditioned function.
The second experiment revealed that ARCH is overwhelmingly more efficient than the
\del{existing}{} CHTs that do not exploit the explicit constraints. This
\del{indicates}{}\new{implies} that the explicit constraints should be exploited
inin the constraint handling, which ARCH effectively achieves. \new{However, note
that the CHTs used in the comparison are limited to those available to the
CMA-ES. Further comparison with state-of-the-art methods is necessary to show the
usefulness of ARCH among existing approaches.}
On the other hand, the comparison to the active-set (1+1)-ES showed a possibility
to further improve the efficacy of ARCH in terms of the number of the objective
function calls. Incorporating the idea of the active-set ES into the repair
operator in ARCH is a possible direction of a future work.

ARCH is designed to solve the constrained continuous optimization problems where the
constraints are all explicit. Box-constrained optimization problems are such examples,
where ARCH exhibits superior performance to an existing box constraint handling
technique. However, it is often the case that there exist explicit constraints and
simulation-based constraints at the same time, where the explicit constraints are
prerequisites to a simulator that computes the objective function value and the
simulation-based constraint violation values. ARCH is expected to be inefficient for
such problems as it requires a substantial amount of constraint evaluations. To
deal with the combination of these two types of constraints, one might need to combine
an explicit constraint handling such as ARCH and a simulation-based constraint
handling such as the multiple constraint ranking (MCR) technique. The invariance
properties of ARCH will play an important role to preserve the invariance properties
of the entire search algorithm when two CHTs are combined. This is an important
direction of future work.

\subsubsection*{Acknowledgements}
This work was supported by the JSPS KAKENHI under Grant Numbers 19H04179 and
19J21892.

\begin{appendices}

\section{Proofs}

\subsection{Proof of \Cref{theorem:affine}}
\label{proof:affine}
Let $\pp_x \in V$ and $\pp_y \in V$ be the vectors of which the coordinates in $\psi$ are $\xx \in \R^n$ and $\yy \in \R^n$, respectively. Firstly, observe
\begin{align*}
  \textstyle \norm{\pp_x - \pp_y}_{\Sigma_V^{-1}}^2
  &= \textstyle \inner{\pp_x - \pp_y}{\Sigma_V^{-1}(\pp_x - \pp_y)} \\
  &= \textstyle \inner{\sum_{i=1}^{n}[\xx - \yy]_i \bm{e}_i}{ \Sigma_V^{-1} ( \sum_{i=1}^{n}[\xx - \yy]_i \bm{e}_i )} \\
  &= \textstyle \inner{\sum_{i=1}^{n}[\xx - \yy]_i \bm{e}_i}{ \sum_{i=1}^{n}[\xx - \yy]_i \Sigma_V^{-1} ( \bm{e}_i )} \\
  &= \textstyle \sum_{i=1}^{n} \sum_{j=1}^{n} [\xx - \yy]_i [\xx - \yy]_j \inner{\bm{e}_i}{ \Sigma_V^{-1} ( \bm{e}_j )} \\
  &= \textstyle (\xx - \yy)^\mathrm{T} \Sigma_\psi^{-1} (\xx - \yy)
    = \textstyle \norm{\xx - \yy}_{\Sigma_\psi^{-1}}^2 \enspace.
\end{align*}
Moreover, it is obvious that $\mathcal{S} = \psi(\mathcal{S}^V)$,
$\J(\xx) = \J^V(\pp_x)$, and $\mathcal{A}(\xx) = \psi(\mathcal{A}^V(\pp_x))$.
Hence, the solution $\tilde{\xx} = \mathtt{Repair}(\xx)$ and the solution $\tilde{\pp}_{x} = \mathtt{Repair}^{V}(\pp_{x})$ are equivalent; that is,
$\tilde{\xx} =  \psi(\tilde{\pp}_x)$.
Then, we also have
$f(\tilde{\xx}) = f^V(\tilde{\pp}_x)$
and $g_\Sigma(\xx) = g_\Sigma^{V}(\pp_x)$, leading to $\Rf(\xx) = \Rf^V(\pp_x)$
and $\Rg(\xx) = \Rg^V(\pp_x)$. As
$\norm{\mu_\psi - \mu_\psi^\mathrm{feas}}_{\Sigma_\psi^{-1}} = \norm{\mu_V - \mu_V^\mathrm{feas}}_{\Sigma_V^{-1}}$
$\norm{\mu_\psi - \mathtt{Repair}(\mu_\psi)}_{\Sigma_\psi^{-1}} = \norm{\mu_V - \mathtt{Repair}^{V}(\mu_V)}_{\Sigma_V^{-1}}$
, it is easy to observe that $\dm^{(t+1)}$ computed in \textsc{ARCH} and $\textsc{ARCH}^V$ are equal under the same $\dm^{(t)} = \dm$. This also leads to the same $\alpha^{(t+1)}$ computed in \textsc{ARCH} and $\textsc{ARCH}^V$ under the same $\alpha^{(t)} = \alpha$. Finally, we obtain $\RT(\xx) = \RT^V(\pp_x)$, which demonstrates that the outputs of the operations of \textsc{ARCH} and $\textsc{ARCH}^V$ are the same. This completes the proof.

\subsection{Proof of \Cref{theorem:increasing}}
\label{proof:increasing}
Firstly, we observe that $g_j(\xx) \leq 0$ if and only if $h_j(g_j(\xx)) \leq 0$ for any $\xx \in \R^n$ and any $j = 1, \dots, m$.
Therefore, it is clear that the feasible domain $\mathcal{S}$,
the set $\J(\xx)$ of indices of the violated constraints,
and the intersection $\mathcal{A}(\xx)$ of the violated constraints are unchanged by any $H$.
As the Mahalanobis distance is not affected by $H$, the above facts imply that the repair
operation remains unchanged; that is, $\mathtt{Repair}(\xx)$ for any $\xx$ is the same on $F$ and $H \circ F$,
and hence, $g_\Sigma(\xx)$ is unchanged.
This implies that $\dm$, and hence also $\alpha$, are unchanged.

Moreover, the rankings $\Rf$ and $\Rg$ are not affected by $H$ because
$\1{f(\yy) < f(\xx)}$ and $\1{f(\yy) = f(\xx)}$ are equivalent to
$\1{h_0(f(\yy)) < h_0(f(\xx))}$, and $\1{h_0(f(\yy)) = h_0(f(\xx))}$
and $g_\Sigma(\xx)$ is unchanged by $H$. Hence, $\RT$ is not
affected by $H$.

Therefore, all of the outputs of \textsc{ARCH} remain unchanged by any $H$.

\section{Implementation remarks for the repair operation}
\label{app:implementation}

In practice, numerical errors in the implementation of the repair operator \cref{eq:intersection,eq:nearest} must be dealt with as numerical optimization routines that sometimes return solutions that violate the constraints slightly.
To guarantee the production of a feasible solution so that it can be evaluated on $f$, we replace all of the constraints in \eq{eq:intersection} and \eq{eq:nearest} with $\g(\xx) \preceq - \epsilon^\ite{t}$.
Then, even with numerical errors, the repair operator is likely to return a feasible solution (that is, $\g(\mathtt{Repair}(\xx)) \preceq 0$) if it is solvable.
We state that the repair operation is successful if a repaired solution is feasible.

The repair operator is implemented as follows:
Firstly, we attempt to solve \eq{eq:intersection} and return the solution if it is feasible.
Otherwise, we attempt to solve \eq{eq:nearest} and return the solution if it is feasible.
If a repaired candidate is still infeasible, we cannot evaluate its $f$-value.
Instead, we set an artificial value, $f^\mathrm{infeas}$, which is treated as $f^\mathrm{infeas} > f(\xx)$ for all $\xx \in \mathcal{S}$ and $f^\mathrm{infeas} = f(\xx)$ for all $\xx \notin \mathcal{S}$.
The $f$-ranking \eq{eq:rf} is computed using $f^\mathrm{infeas}$, i.e., the
worst $f$-ranking is assigned to such a point, while the computation of the $g$-ranking \eq{eq:rg} remains unchanged: we use the Mahalanobis distance between the original candidate $\xx$ and unsuccessfully repaired candidate $\mathtt{Repair}(\xx)$.
To maintain a high probability of success of the repaired operation, we adapt $\epsilon^\ite{t}$.
We set $\epsilon^\ite{0} = 10^{-13}$ and update it as
\begin{align*}
  \epsilon^\ite{t+1} = \epsilon^\ite{t} \times
  \begin{cases}
    \frac{1}{2} & |\{\text{unsuccessfully repaired points at iteration $t$}\}| \leq \lceil 0.1\lambda \rceil \enspace, \\
    10          & \text{otherwise} \enspace,
  \end{cases}
\end{align*}%
and $\epsilon^\ite{t+1}$ is clipped to $[10^{-15}, 10^{-4}]$.

The optimization routine for the repair operation can be selected according to the problem.
For example, if the constraints are linear and not redundant, we can obtain the repaired point without relying on gradient-based minimization of \eqref{eq:intersection}.
If nonlinear constraints are considered, we employ SLSQP
\citep{kraft1988software}. Whether or not the optimization problems \eqref{eq:intersection}
and \eqref{eq:nearest} are solvable depends on the assumptions
on the internal optimization method. If these assumptions are not satisfied, the repair operation is not guaranteed to return the optimal solution, and it may output infeasible or feasible but sub-optimal solutions.

\section{The reason that \cref{eq:intersection} is preferred over \cref{eq:nearest}}
\label{app:repair}

\begin{figure}[t]
  \centering
  \begin{subfigure}{0.5\hsize}
    \includegraphics[width=\subsinglefigsize]{Ranking_Sphere20_Box_seed100.pdf}%
  \end{subfigure}%
  \begin{subfigure}{0.5\hsize}
    \includegraphics[width=\hsize]{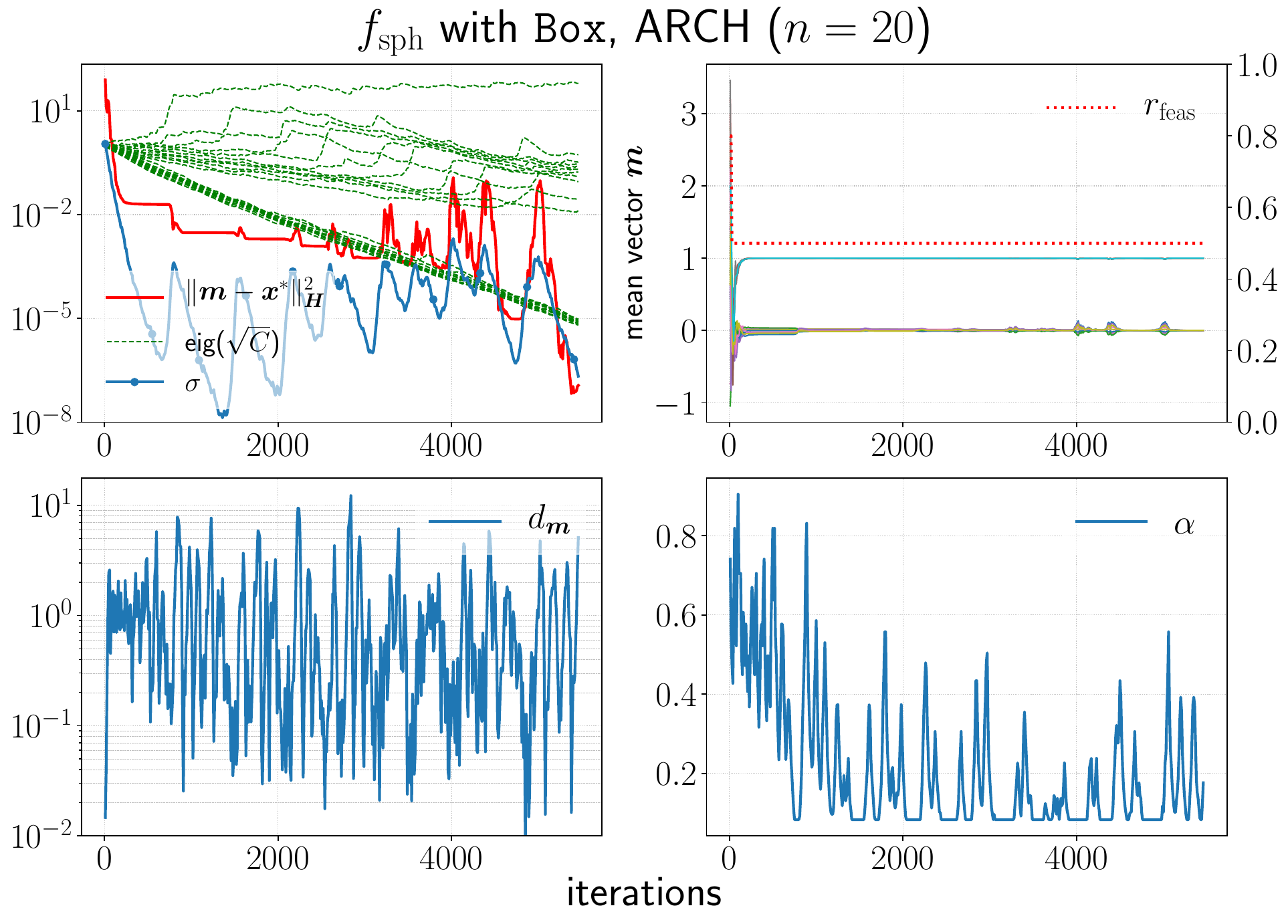}
  \end{subfigure}%
  \caption{Behavior of ARCH solving \cref{eq:intersection}(left) and \cref{eq:nearest}(right) for
    repair.}\label{fig:nearest}%
\end{figure}

We describe why \cref{eq:intersection} is preferable to \cref{eq:nearest}.
\Cref{fig:nearest} illustrates a typical run on $\fsph$ with $\BOX$%
, where the experiment is the same as \Cref{sec:exp1}.
We observe that the eigenvalues of the sampling distribution, namely $\text{eig}(\sqrt{\cov})$, are divided into two groups.
This is similar to the behavior of the resampling in \Cref{fig:single}.
The reason for the behavior is described as follows.
The points repaired on the intersection of the boundaries of the constraints that are active at the optimum tend to exhibit better $f$-values than the other points.
We refer to the region where the points are repaired onto the abovementioned intersection as a preferred region of the algorithm.
Once solutions in the preferred region have been determined,
they are ranked highly, and the next covariance matrix is updated to increase the likelihood of these solutions.
This tends to increase the eigenvalues in the directions to the preferred region.
The preferred region is dependent on the distribution shape, and it will become sharper as the distribution does, resulting in a very sharp distribution.
The step-size first dropped to a very small value as it does on very
  ill-conditioned functions and increased once the covariance matrix learned the
  long axes.
The preferred region of the algorithm using \eqref{eq:intersection} is less dependent on the distribution shape,
and performs significantly better than that using \eqref{eq:nearest}.
This is why we attempt to solve \cref{eq:intersection} first.

\begin{figure}[t]
  \centering
  \begin{subfigure}{\medfigsize}
    \includegraphics[width=\submedfigsize]{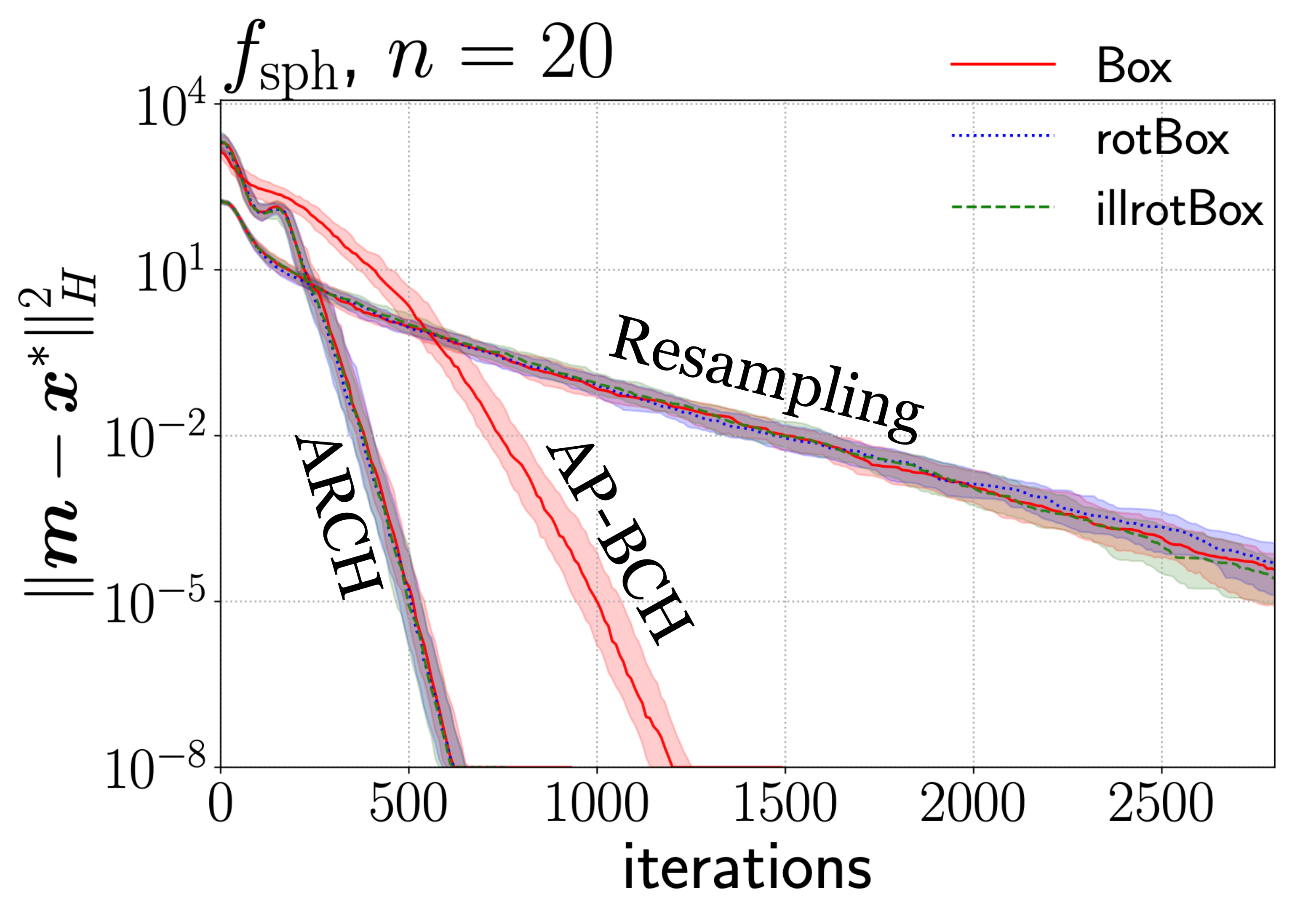}
  \end{subfigure}%
  \begin{subfigure}{\medfigsize}
    \includegraphics[width=\submedfigsize]{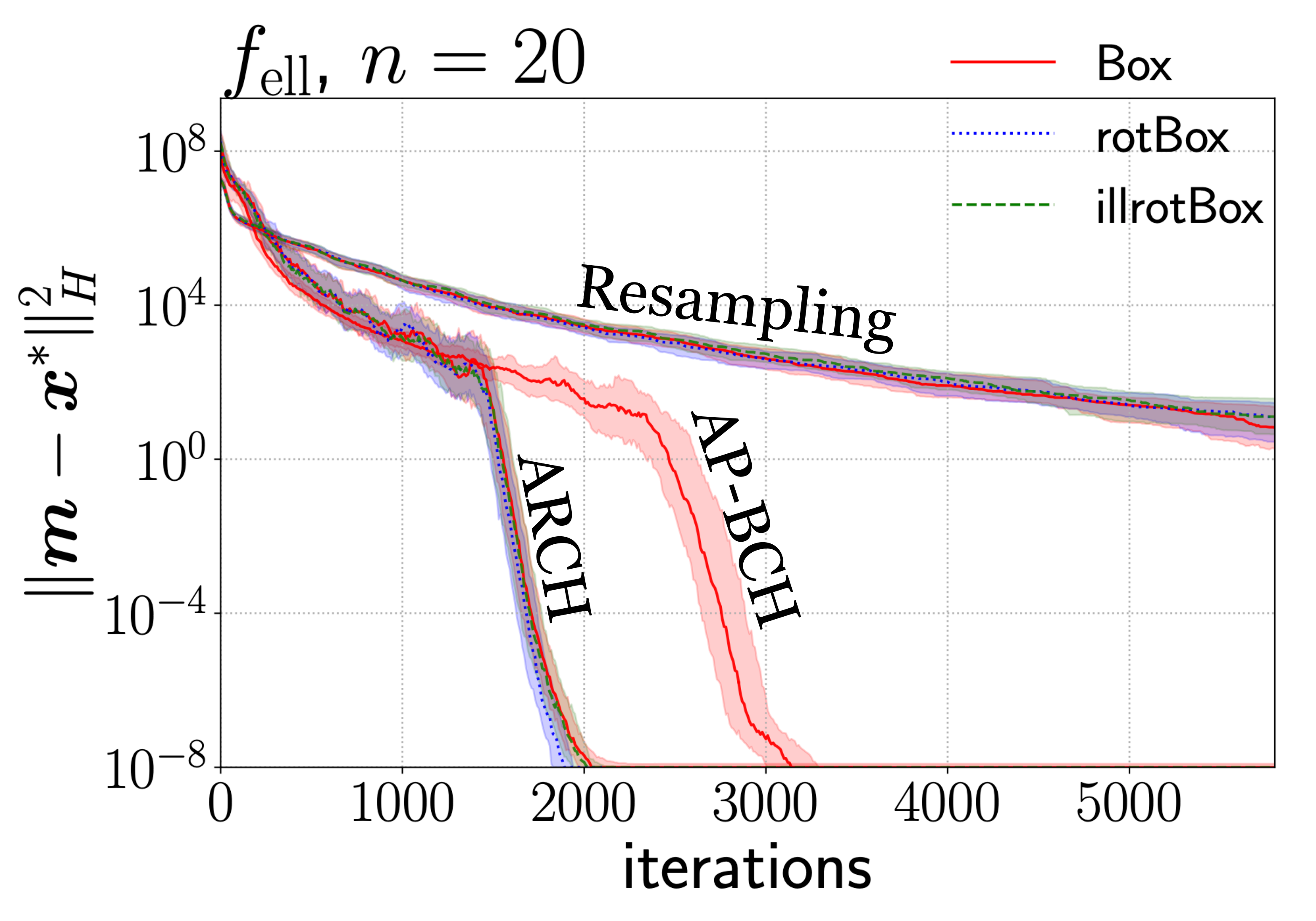}
  \end{subfigure}%
  \begin{subfigure}{\medfigsize}
    \includegraphics[width=\submedfigsize]{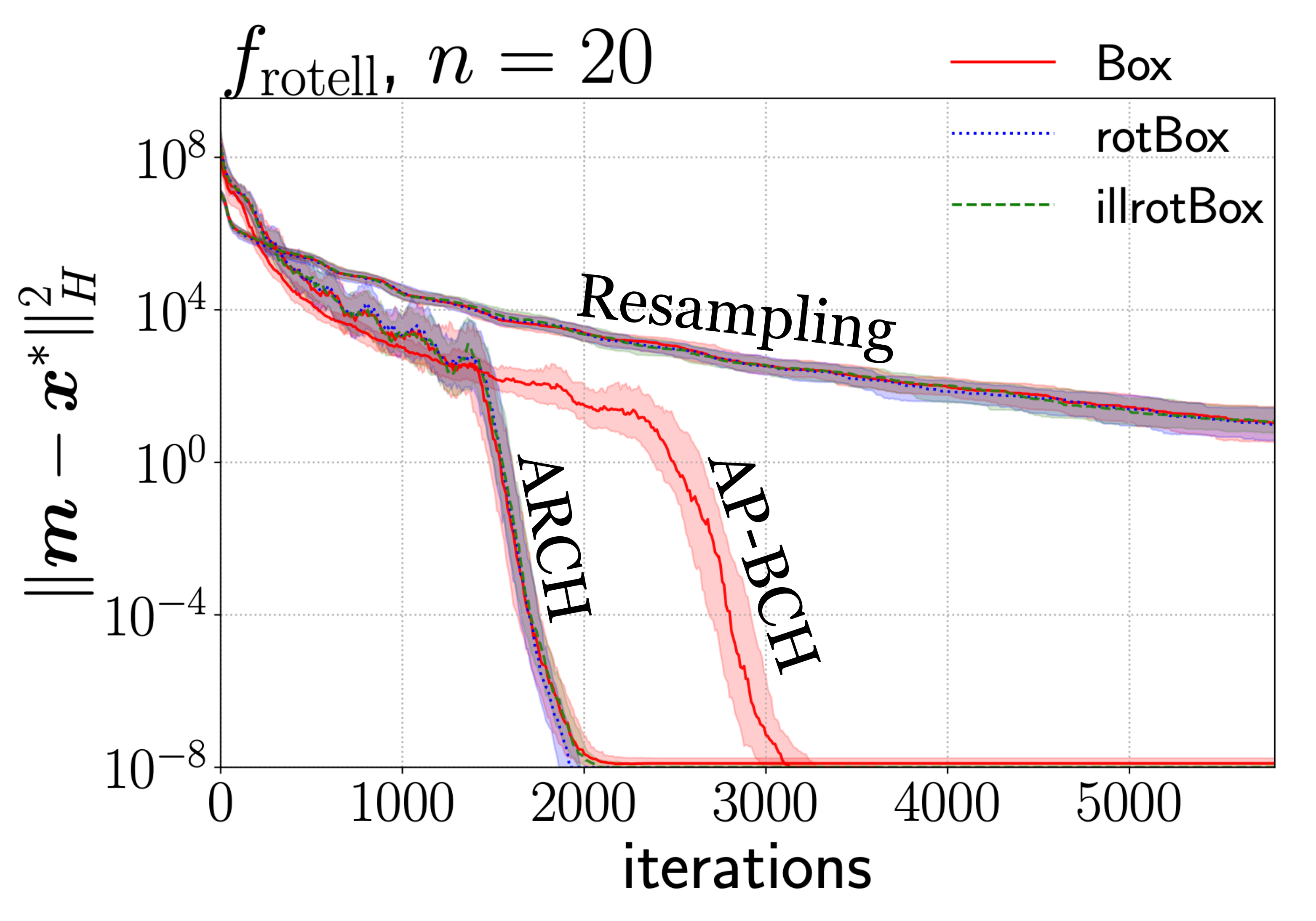}
  \end{subfigure}%
  \\
  \begin{subfigure}{\medfigsize}
    \includegraphics[width=\submedfigsize]{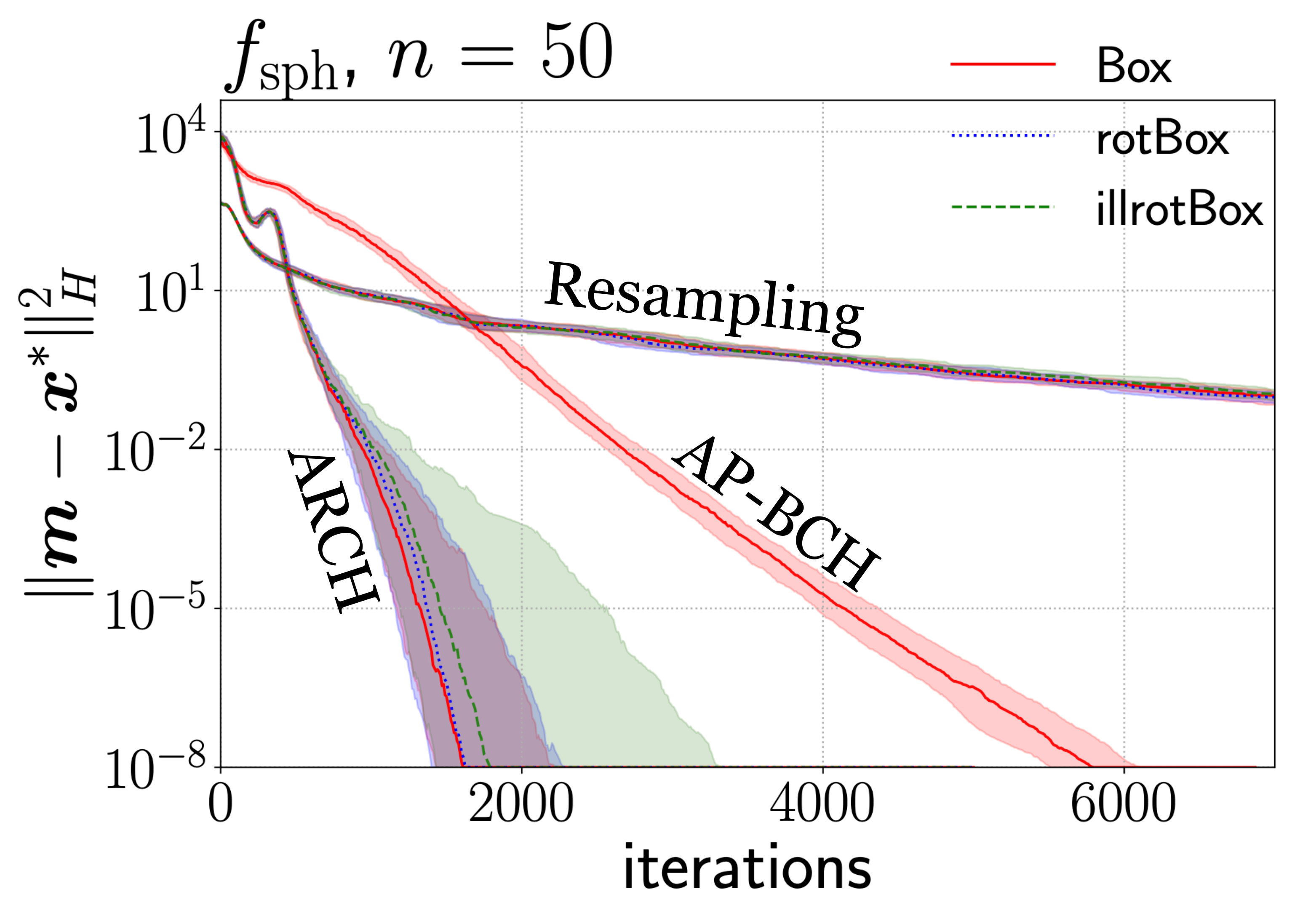}
  \end{subfigure}%
  \begin{subfigure}{\medfigsize}
    \includegraphics[width=\submedfigsize]{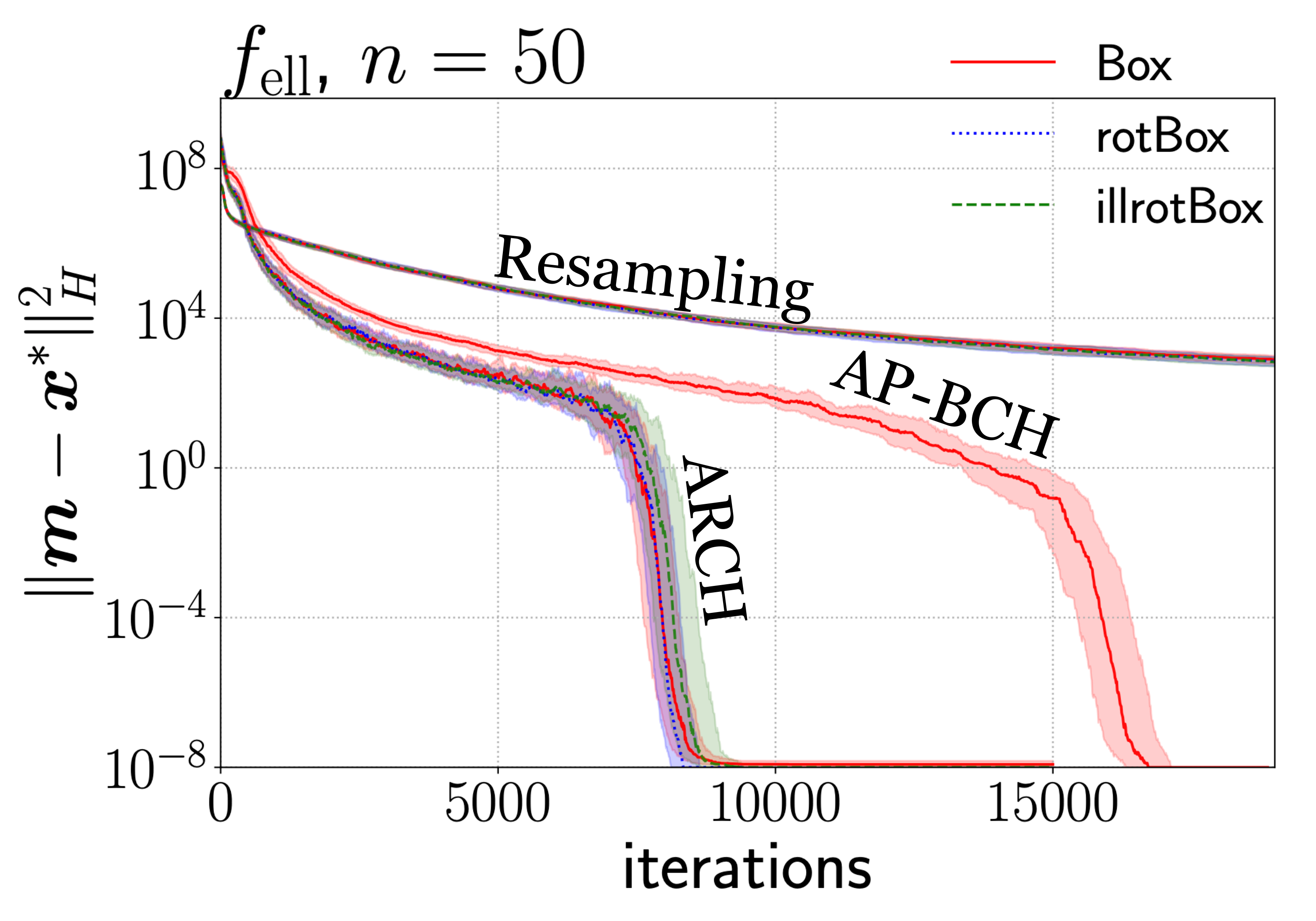}
  \end{subfigure}%
  \begin{subfigure}{\medfigsize}
    \includegraphics[width=\submedfigsize]{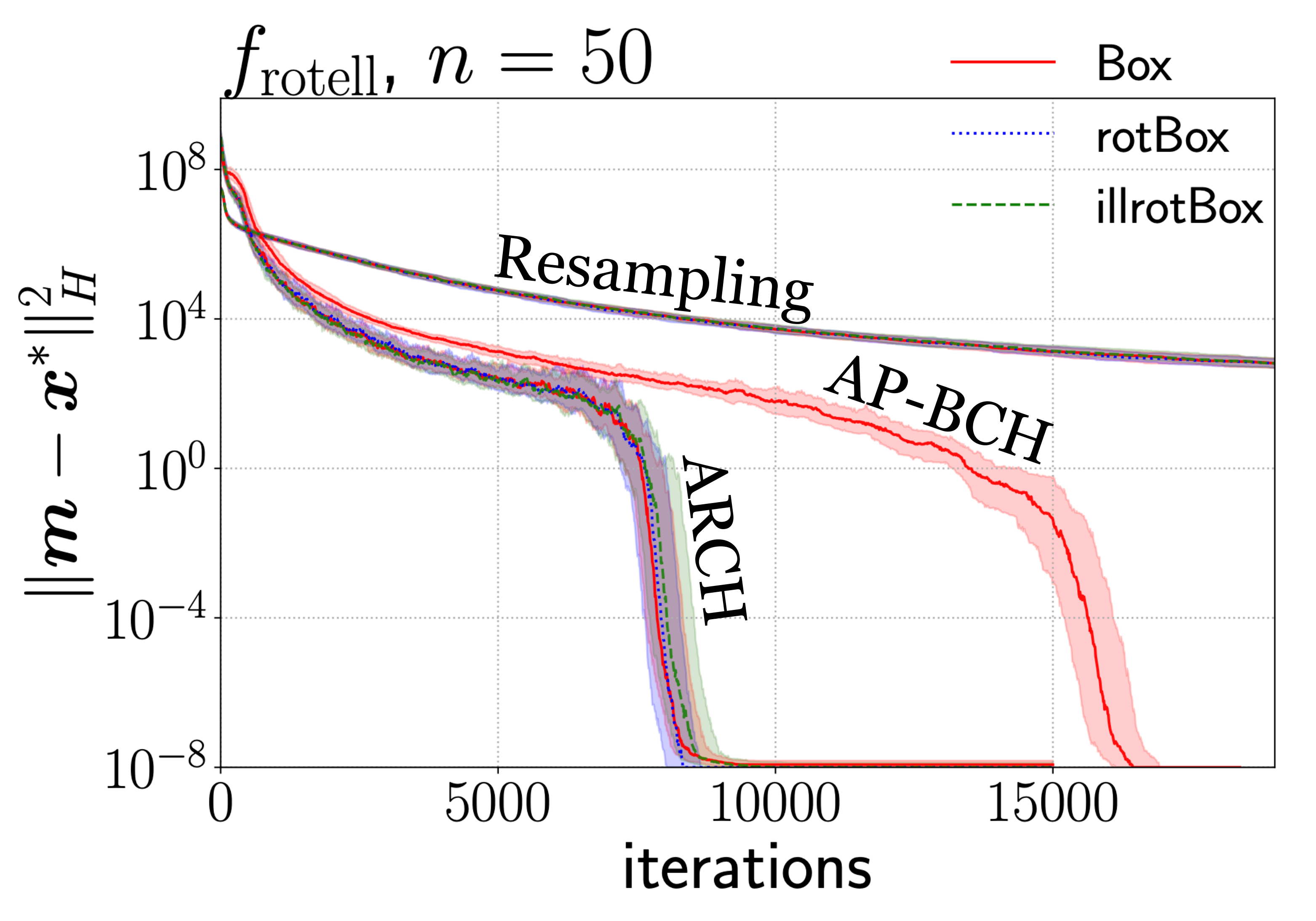}
  \end{subfigure}
  \caption{Median (line) and 25\% to 75\%-ile range (band) over 100 trials.}
  \label{fig:med-general}
\end{figure}

\new{
\section{Additional Experiments on Linearly Constrained Problems}
\label{sec:lin-prob-general}

We show additional experimental results on linearly constrained
problems.
In \Cref{sec:exp1}, the initial covariance matrix
was set as $\cov^{(0)}_{\BOX} = \I$ on the $\BOX$ problem, $\cov^{(0)}_{\rotBox} =
\bm{P}^{-1}_{\rm rot} \cov^{(0)}_{\BOX} (\bm{P}^{-1}_{\rm rot})^\T$ on the
$\rotBox$ problem, and $\cov^{(0)}_{\illrotBox} = \bm{P}^{-1}_{\rm illrot}
\cov^{(0)}_{\BOX} (\bm{P}^{-1}_{\rm illrot})^\T$ on the $\illrotBox$
problem.
That is, the shape of the initial search distribution matches
the shape of the $n$-parallelotope-shaped feasible domain.
In this experiment, we mismatch the initial search distribution
with the shape of the feasible domain by setting
$\cov^{(0)}_{\illrotBox} = \I$ on the $\illrotBox$ problem,
i.e., $\cov^{(0)}_{\BOX} = \bm{P}_{\rm illrot} (\bm{P}_{\rm illrot})^T$ on the
$\BOX$ problem.
Experimental settings are the same as in \Cref{sec:exp1}, except for the initial
covariance matrix $\cov^{(0)}$.

The results are shown in \Cref{fig:med-general}.
The optimization progress is measured by
the Mahalanobis distance between the mean vector and optimal solution $\norm{\mm -
  \xx^*}^2_{\HH}$ given the Hessian matrix $\HH$ of the objective function.

The lines of ARCH overlap as in \Cref{fig:medfig} except for deviations observed on
$f_\text{sph}$ with $n = 50$, which is due to numerical errors of the repair operation.
Comparing
the convergence speeds on $\fsph$, ARCH reached the target threshold with less than the half of iterations spent by AP-BCH,
while it was almost the same in \Cref{fig:medfig}. The difference is due to the
adaptation speed of the covariance matrix. In this experiment, the initial
covariance matrix is ill-conditioned and rotated, i.e., it is necessary to
learn the isotropic scale of the sphere function.
}%

\end{appendices}


\end{document}